\documentclass[10pt,twocolumn,letterpaper]{article}

\usepackage{cvpr}              
\definecolor{cvprblue}{rgb}{0.21,0.49,0.74}
\usepackage[pagebackref,breaklinks,colorlinks,allcolors=cvprblue]{hyperref}

\usepackage{url}
\usepackage{graphicx}
\usepackage{amsmath}
\usepackage{amssymb}
\usepackage{mathtools}
\usepackage{amsthm}
\usepackage{enumerate}
\usepackage{comment}
\usepackage{enumitem}

\usepackage{bbm}
\usepackage{arydshln}
\usepackage{bm}

\usepackage{tabularx}
\usepackage{multirow}
\usepackage{tabularray}
\usepackage{tabulary}
\usepackage{tabularx}
\usepackage{makecell}
\usepackage{balance}
\usepackage{soul}
\usepackage{wrapfig}
\usepackage{lipsum}
\usepackage{float}
\usepackage{lscape}
\usepackage[toc,page,header]{appendix}
\usepackage{xspace}
\usepackage{needspace}
\usepackage{blindtext}
\usepackage[table]{xcolor}

\usepackage{enumitem}
\usepackage{comment}
\usepackage{courier}
\usepackage{booktabs}
\usepackage{caption}
\usepackage{amsthm} 
\theoremstyle{plain}
\newtheorem{theorem}{Theorem}[section]

\theoremstyle{definition}
\newtheorem{definition}[theorem]{Definition}

\theoremstyle{remark}

\usepackage{algorithm}
\usepackage{algpseudocode}

\algtext*{EndIf}
\algtext*{EndFor}
\algtext*{EndWhile}
\algtext*{EndFunction}
\algtext*{EndProcedure}

\def\paperID{} 
\def\confName{CVPR}
\def\confYear{2026}

\title{See and Fix the Flaws: Enabling VLMs and Diffusion Models \\ to Comprehend Visual Artifacts via Agentic Data Synthesis}

\newcommand{\myparagraph}[1]{\vspace{0.15cm}\noindent\textbf{#1}}

\newcommand{\algname}{{ArtiAgent}} 
\newcommand{\tblalgname}{\textbf{ArtiAgent}}
\newcommand{\algdescrp}{a novel agentic framework that synthesizes artifacts for
arbitrary visual contexts without human intervention}

\newcommand{\benchname}{ArtiBench}

\author{
Jaehyun Park$^{1,*}$\hspace{0.5em}Minyoung Ahn$^{2,3,*}$\hspace{0.5em}Minkyu Kim$^{3}$\hspace{0.5em}Jonghyun Lee$^{3}$\hspace{0.5em}Jae-Gil Lee$^{1}\hspace{0.5em}$Dongmin Park$^{3,\dagger}$\\
$^{1}$KAIST\quad$^{2}$Seoul National University\quad$^{3}$KRAFTON \\
{\tt\footnotesize\{jhpark813,jaegil\}@kaist.ac.kr,\{minkyu.kim,jonghyunlee,dongmin.park\}@krafton.com,michellahn02@snu.ac.kr}
\vspace{-0.35cm}
}

\begin{document}
\maketitle
\begingroup
\renewcommand\thefootnote{}
\footnotetext{* Equal contribution; work done during an internship at KRAFTON.}
\footnotetext{$\dagger$ Corresponding author.}
\endgroup

\begin{abstract}
    Despite recent advances in diffusion models, AI generated images still often contain visual artifacts that compromise realism. Although more thorough pre-training and bigger models might reduce artifacts, there is no assurance that they can be completely eliminated, which makes artifact mitigation a highly crucial area of study.  
    Previous artifact-aware methodologies depend on human-labeled artifact datasets, which are costly and difficult to scale, underscoring the need for an automated approach to reliably acquire artifact-annotated datasets. In this paper, we propose \textbf{\algname{}}, which efficiently creates pairs of real and artifact-injected images. It comprises three agents: a perception agent that recognizes and grounds entities and subentities from real images, a synthesis agent that introduces artifacts via artifact injection tools through novel patch-wise embedding manipulation within a diffusion transformer, and a curation agent that filters the synthesized artifacts and generates both local and global explanations for each instance. Using \algname{}, we synthesize 100K images with rich artifact annotations and demonstrate both efficacy and versatility across diverse applications. Code is available at \href{https://github.com/krafton-ai/Artiagent}{link}.
\end{abstract}
\vspace*{-0.7cm}
\section{Introduction}
\label{sec:intro}

\begin{figure*}[t]
    \centering
    \vspace{-0.3cm}
    \includegraphics[width=\textwidth]{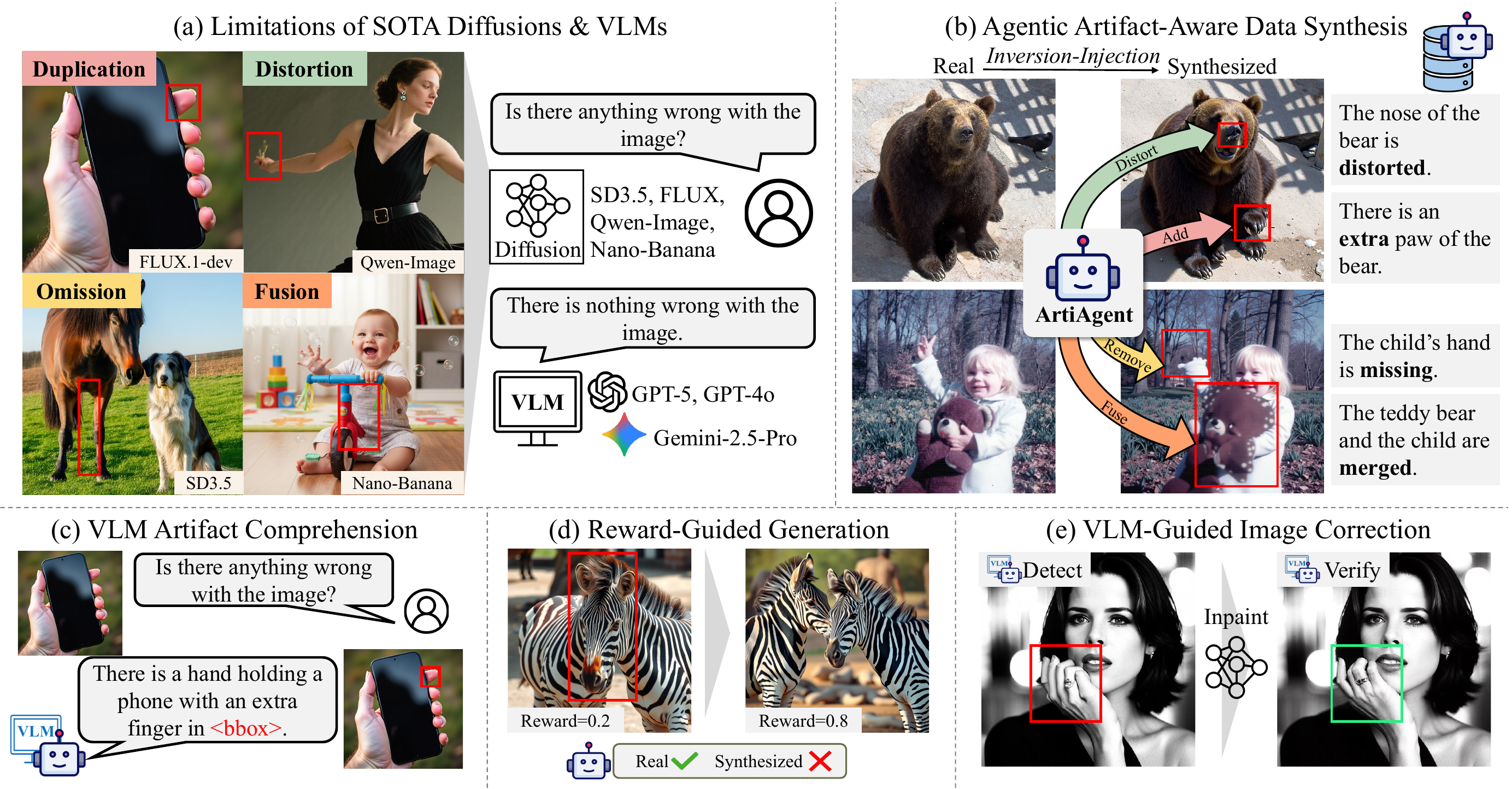}
    \vspace{-0.7cm}
    \caption{Overview of our challenges and approach. The red boxes indicate the regions with visual artifacts. (a) Examples of structural visual artifacts in state-of-the-art diffusion models and the inability of VLMs to recognize or explain them. (b) Overview of \algname{}, \algdescrp{}. (c) Example of VLM-based artifact comprehension via detection, explanation, and localization. (d) Application to reward-guided text-to-image generation. (e) Application to image correction, where artifact-aware VLM-guided inpainting removes the flawed regions.}
    \vspace{-0.6cm}
    \label{fig:fig1}
\end{figure*}

Recent advances in diffusion models have led to remarkable success in generating highly photorealistic images~\citep{zhang2023text, saharia2022photorealistic}. 
While these models have demonstrated striking achievements in text-to-image alignment and aesthetic quality, they still suffer from producing \emph{visual artifacts}, unintended distortions or anomalies in generated outputs. 
For example, in Figure~\ref{fig:fig1}(a), even state-of-the-art models such as Nano-Banana~\citep{google2025nanobanana} produce artifacts, e.g., \emph{six-fingered hands} and \emph{fused entities}, reducing user satisfaction, e.g., uncanny valley.
Furthermore, mitigating such artifacts is especially critical in high-stakes diffusion applications where reliability is paramount, including image generation in medicine~\citep{khader2023medical}, robotics~\citep{huang2025robotics}, autonomous driving~\citep{jiang2024scenediffuser}, and so on.

Meanwhile, in image understanding, vision-language models (VLMs) have made substantial progress, showing strong capabilities in visual question answering and scene description~\citep{zhang2024vision}. These advances have allowed VLMs to serve as automatic systems for many vision tasks, including medical analysis~\citep{hartsock2024vision} and robotics control~\citep{kim2024openvla}. However, we find that VLMs face challenges when confronted with visual artifacts. As shown in Figure~\ref{fig:fig1}(a), even state-of-the-art VLMs, such as GPT-5~\cite{openai2025gpt5} and Gemini-2.5-pro~\citep{comanici2025gemini}, exhibit limited ability to detect, localize, or explain artifacts in AI-generated images, nearly indistinguishable from random guessing (see $\S$~\ref{sec:exp} for details). Consequently, VLMs cannot yet serve as reliable systems for artifact comprehension, which constrains their utility.

Recently, several approaches have been proposed to tackle visual artifacts in VLMs and diffusions\citep{zhang2023pal, kang2025legion, wang2025diffdoctor}. However, existing efforts exhibit two major limitations: (1) they mainly target simple artifact types (e.g., Gaussian noise or blur), which were prevalent in early diffusion models such as SD1.0~\citep{rombach2022ldm}, but are rarely observed in modern models, where more \emph{plausible} physical distortions are predominant; and (2) they rely heavily on human-annotated artifacts (e.g., as many as 10K labels), which is costly and fundamentally limited in scalability to capture the full diversity of diffusion-generated artifacts. These limitations highlight the need for scalable, annotator-free methods to address plausible and extensive artifacts across diverse visual contexts produced by modern diffusion models.

To fill this gap, we introduce \textbf{\algname{}} (Figure~\ref{fig:fig1}(b)), a novel agentic framework that injects plausible artifacts without human intervention via inversion-restoration~\citep{deng2025fireflow} by perturbing the attention of DiT (see $\S$~\ref{subsubsec:inversion_injection}). \algname{} consists of three agents: (1) a \emph{perception agent}, which identifies the most suitable objects or entities in a real image to perturb, (2) a \emph{synthesis agent}, which selects and applies the tools to generate plausible artifacts, and (3) a \emph{curation agent}, which filters low-quality results and refines their annotations. Through this agentic pipeline, \algname{} produces high-quality artifact-injected images with rich annotations, including binary labels, locations, and explanations suitable for detection, localization, and reasoning.

For thorough validation, we present \textbf{\benchname{}}, a new benchmark of 1K AI-generated images produced by modern generative models such as FLUX-dev and Nano-Banana, where each image is carefully annotated by humans with binary artifact labels, their bounding box locations, and explanations. We evaluate \algname{} on \benchname{} as well as three existing benchmarks, showing that open-source VLMs (e.g., Qwen2.5-VL) fine-tuned on 100K training samples generated by \algname{} consistently outperform proprietary VLMs (e.g., GPT-5) and prior baselines across major artifact-perception tasks, including detection, localization, and reasoning. Furthermore, with these artifact-aware VLM, we benefit two downstream applications: (1) guiding diffusion sampling toward artifact-free generation with VLM-based artifact reward, and (2) automatically editing diffusion outputs that contain artifacts, substantially improving the image generation pipelines.

Our key contributions are summarized as follows:
\begin{itemize}[leftmargin=10pt,noitemsep]
\item \textbf{Framework.} We introduce \algname{}, an agentic data synthesis framework that produces diverse plausible artifacts at scale, enriched with high-quality annotations.
\item \textbf{Tools.} We develop agentic tools that inject artifacts via our novel inversion-injection method during image reconstruction, which can be used by any DiT model.
\item \textbf{Datasets.} We synthesize a large-scale training set with 100K examples via \algname{}, along with \benchname{}\footnote{{\scriptsize\url{https://huggingface.co/datasets/KRAFTON/ArtiBench}}}, a challenging benchmark of 1K images generated by modern generative models with human labels.
\item \textbf{Experiments.} Extensive experiments demonstrate the superiority of \algname{} by scaling up VLM performance on core artifact-perception tasks. Moreover, we show its utility in downstream diffusion-based applications: artifact-free image generation and editing.
\end{itemize}
\section{Related Work}
\label{sec:related_work}


\myparagraph{Visual Artifact Datasets.}
Several training datasets have been introduced to supervise the understanding of visual artifacts in generative models.
PAL\,\citep{zhang2023pal} provides 10K images with pixel-level annotations of perceptual defects for segmentation-based training.
SynthScars\,\citep{kang2025legion} provides 12K images with pixel-level masks and textual explanations.
DiffDoctor\,\citep{wang2025diffdoctor} scales human labeling through semi-supervised expansion from a 25K seed set. Although these datasets offer essential supervision for artifact detection and correction, their reliance on human annotation makes them costly and difficult to scale.

Moreover, several benchmark datasets have been released to evaluate models' ability to understand visual artifacts.
RichHF-18K\,\citep{liang2024rich} provides annotated artifact regions, while LOKI\,\citep{ye2024loki} adds natural-language explanations with bounding box labels.
SynthScars also releases evaluation splits to test perception, localization, and explanation capabilities.
These benchmarks serve as reference points for measuring a model's artifact awareness; however, most of them rely on artifacts from earlier diffusion models and tend to focus on degenerate artifacts (e.g., Gaussian noise), limiting their relevance to the richer and more diverse failure modes of modern generative systems.

\myparagraph{Handling Visual Artifacts.} 
Using these datasets, different modeling strategies have been explored. PAL~\citep{zhang2023pal} trains segmentation models to localize artifact regions and enables automated correction via inpainting. RichHF-18K~\citep{liang2024rich} is used to train multimodal models that predict human-like feedback heatmaps, which can then refine diffusion models through preference learning. LEGION~\citep{kang2025legion} trains GLaMM~\citep{rasheed2024glamm} with the SynthScars dataset that allows detection, localization, and explanation into a unified model. DiffDoctor~\citep{wang2025diffdoctor} employs its dataset to train an artifact segmentation model, which is used for diffusion fine-tuning to alleviate artifact generation. While these studies emphasize the value of artifact detection and reasoning, they also reveal the necessity of reducing reliance on manual annotation to enable scalable and reliable dataset construction.
\section{Understanding Visual Artifacts}
\label{sec:prelim}
\subsection{Problem Scope}

As generative models have matured, the types of visual artifacts that frequently appear have also changed. Early GANs~\citep{goodfellow2020generative} and U-Net–based diffusions~\citep{rombach2022high} predominantly exhibited naive distortions such as Gaussian-like noise or low-level pixel corruption~\citep{zhang2023pal}. Modern diffusion models, equipped with DiT-based architectures and trained on higher-quality datasets, have largely overcome these naive distortions. However, they still generate plausible \textit{structural} visual artifacts, which are the main focus of this work.

\begin{definition} (Informal) \emph{Structural visual artifacts} refer to defects in which the inherent physical structures of objects are distorted in the generated image. That is, while the contents specified in the prompt are represented, their form violates common-sense plausibility (e.g., a generated dog with two noses). This definition excludes text-to-image misalignments where the prompt content itself is misrepresented (e.g., generating a cat given the prompt of a dog).
\end{definition}

\subsection{Artifact Analysis in Modern Generative Models}

\begin{figure}[t]
  \centering
  \begin{minipage}{0.53\linewidth}
    \centering
    \captionof{table}{Artifact frequency of modern diffusion models.}
    \vspace{-0.3cm}
    \label{tbl:artifact_freq}
    \small
\begin{tabular}{c|c}
\toprule
Generative Model & Freq. \\ 
\midrule
SD3.5-Large & 36\%\\
FLUX-schnell & 28\%\\
Qwen-Image & 17\%\\
FLUX-dev & 12\%\\
Nano-Banana & 5\%\\
\bottomrule
\end{tabular}
  \end{minipage}\hfill
  \begin{minipage}{0.45\linewidth}
    \centering
    \includegraphics[width=0.75\linewidth]{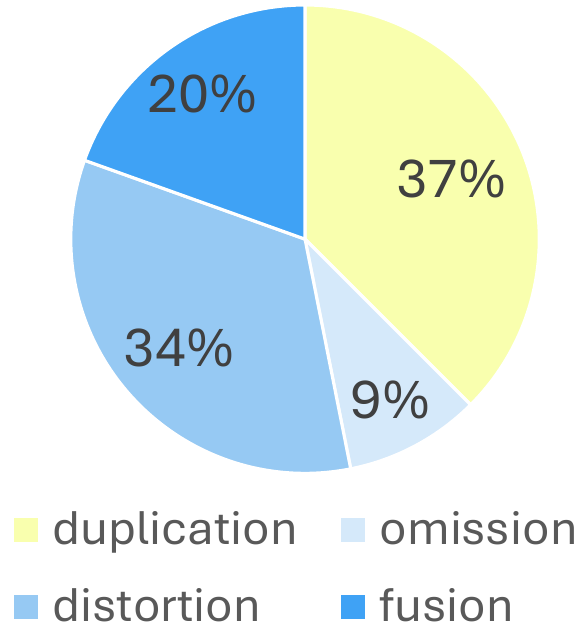}
    \vspace{-0.3cm}
    \captionof{figure}{Artifact type distribution of diffusion models.}
    \label{fig:artifact_dist}
  \end{minipage}
  \vspace{-0.4cm}
\end{figure}


We have categorized structural artifacts into four representative types: \textbf{duplication}, \textbf{omission}, \textbf{distortion}, and \textbf{fusion}. 
Figure~\ref{fig:fig1}(a) provides illustrative examples of each case, showing how diffusion models can compromise reliability while still producing visually high-quality pixels. 
To systematically analyze these artifacts, we randomly sampled 100 captions from the MS-COCO dataset~\citep{chen2015mscoco} and used them as text-to-image prompts across five state-of-the-art diffusion models, including Stable Diffusion 3.5~\citep{esser2024sd3}, FLUX~\citep{FLUX}, Qwen-Image~\citep{wu2025qwenimage}, and Nano-Banana~\citep{google2025nanobanana}. For each generated image, a human annotator manually inspected the outputs and identified whether they contained any artifacts, further categorizing them into the four structural types. 
Based on this evaluation, Table~\ref{tbl:artifact_freq} summarizes the frequency of artifacts for each model, while Figure~\ref{fig:artifact_dist} presents the relative occurrence of different artifact types. 
These findings underscore the importance of addressing structural artifacts in modern generative modeling, highlighting the need for scalable methods to detect, analyze, and mitigate them.

\section{Agentic Pipeline for Artifact Synthesis}
\label{sec:method}

\begin{figure*}[t]
    \centering
    \includegraphics[width=\textwidth]{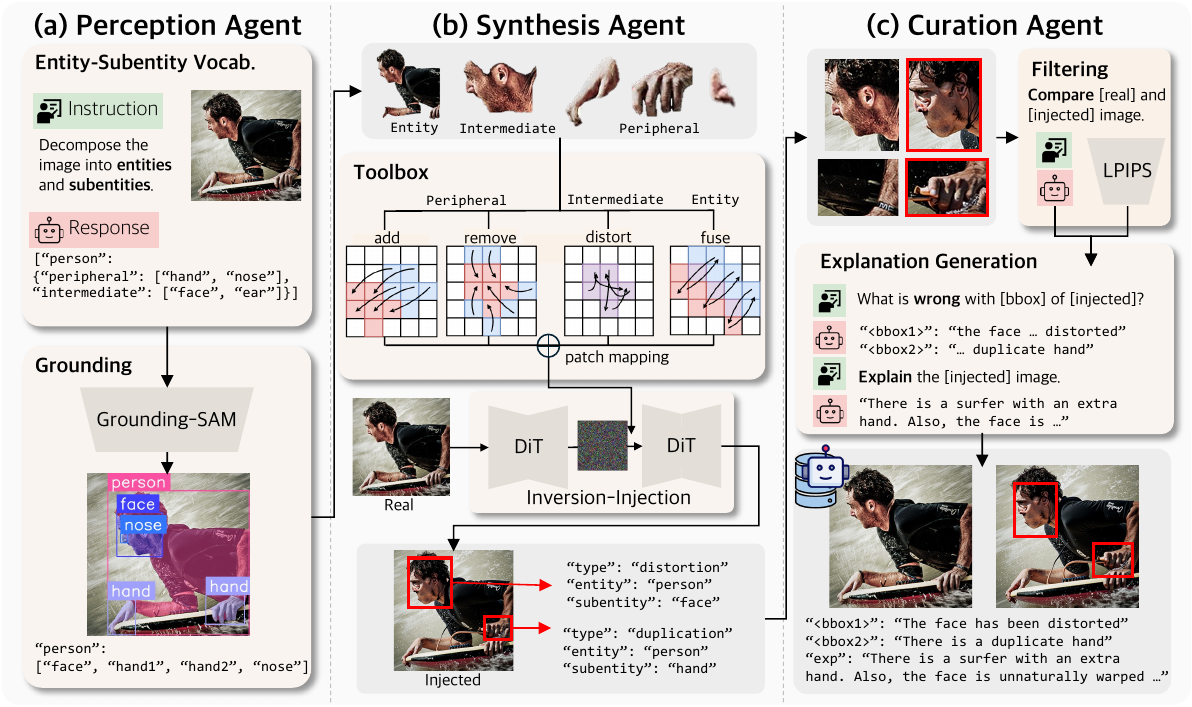}
    \vspace{-0.7cm}
    \caption{\algname{} consists of three coordinated agents: (1) the \emph{perception agent} detects entities and subentities using Grounded-SAM; (2) the \emph{synthesis agent} injects artifacts through patch mapping tool and the inversion-injection paradigm; and (3) the \emph{curation agent} filters low-quality results and generates localized and global textual explanations.}
    \label{fig:method}
    \vspace{-0.2cm}
\end{figure*}

We propose \textbf{\algname{}}, a fully automated agentic pipeline that synthesizes visual artifacts in clean images. As shown in Figure~\ref{fig:method}, three agents (perception, synthesis, curation) select injection candidates, inject artifacts, and curate outputs with local and global explanations. Leveraging recent LLMs and our artifact-injection tools, \algname{} determines the artifact types and locations for each image and produces high-quality annotations. Full agent details and visualizations are in Appendices~\ref{apdx:pipeline}.

\subsection{Perception Agent (Figure \ref{fig:method}(a))}

The perception agent aims to analyze a real image and decompose it into meaningful semantic units that can serve as reliable candidates for the synthesis agent.

\subsubsection{Entity-Subentity Vocabulary Generation}
\label{subsubsec:entity_subentity_vocab}
The agent uses out-of-the-box VLMs to decompose the input image into a hierarchical vocabulary of entities (e.g., dog) and subentities (e.g., nose). These are grouped into two semantic levels: peripheral subentities, such as fingers or legs, and intermediate subentities, such as body or face. The prompt template for this module is in Appendix~\ref{apdx:ent_subent}.

\subsubsection{Entity-Subentity Grounding} The agent then employs Grounded-SAM~\citep{ren2024groundedsam} to ground entities and subentities visible in the image. Once the segmentation masks for both entities (e.g., dog) and subentities (e.g., leg) are obtained, the agent performs a containment analysis to associate each subentity with its parent entity (e.g., a leg of a dog) by computing the overlap ratio.



\subsection{Synthesis Agent (Figure \ref{fig:method}(b))}
\label{subsec:synth_agent}

The synthesis agent uses the perception agent's grounding to inject artifacts into an image via two components: a \textit{toolbox} that generates target-reference patch mappings and an \textit{inversion-injection} module that applies them to the image.

\subsubsection{Target-Reference Patch Mapping Toolbox} 
\label{subsubsec:toolbox}
\begin{figure}[t]
    \centering
    \vspace{-0.5cm}
    \includegraphics[width=\linewidth]{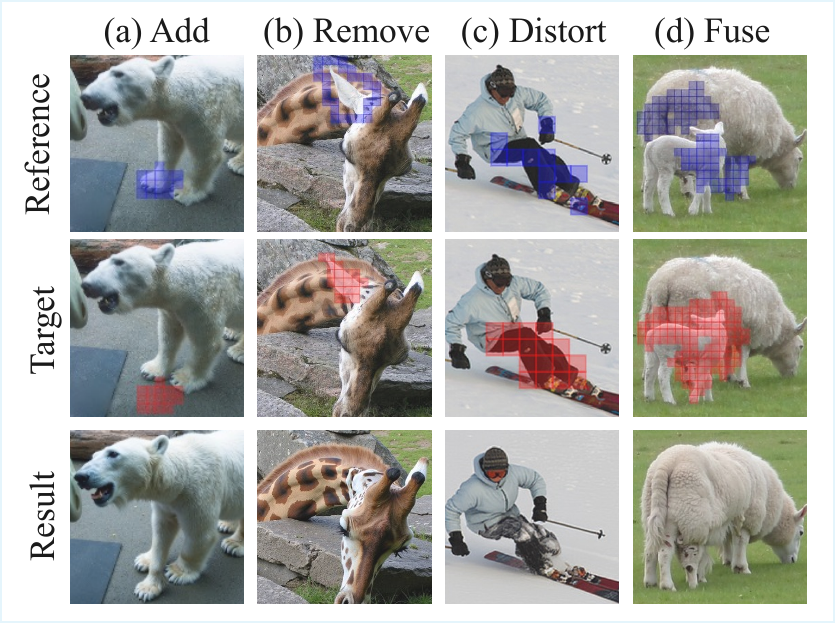}
    \vspace*{-0.9cm}
    \caption{Visualization of each target-reference patch mapping and its resulting artifact-injected image.}
    \label{fig:tool_viz}
    \vspace{-0.6cm}
\end{figure}

The toolbox has four artifact-injection tools: \textbf{add}, \textbf{remove}, \textbf{distort}, and \textbf{fuse}, each producing a target-reference patch mapping. It applies add and remove to peripheral subentities, distort to intermediate subentities, and fuse to two overlapping entities. Figure~\ref{fig:tool_viz} shows one example per tool. Detailed algorithms are in Appendix~\ref{apdx:gen_tools}.

\begin{itemize}[leftmargin=10pt,noitemsep]
\item \textbf{Add.} Reference patches are the original subentity region, while the tool chooses the best target-patch candidate from surrounding patches. Nearby patches without overlaps or subentity class conflicts are preferred.

\item \textbf{Remove.} Target patches are given as the original subentity region, while the tool creates reference patch mappings to replace them with the surrounding context.

\item \textbf{Distort.} Target patches are the original subentity region, while the tool applies kernels to target patches to generate distorted mappings to reference patches. We use three kernels: jitter for random displacement, strip for circular shifts over bands, and a random permutation kernel.

\item \textbf{Fuse.} The fuse tool generates mappings that blend content across multiple entities. The target patches contain the overlapping region between two entity instances, and the chunks of reference patches from one entity are assigned to the target patches of the other entity.

\end{itemize}

\subsubsection{Inversion-Injection Method for Artifact Synthesis}
\label{subsubsec:inversion_injection}

\begin{figure}[t!]
    \centering
    \includegraphics[width=\linewidth]{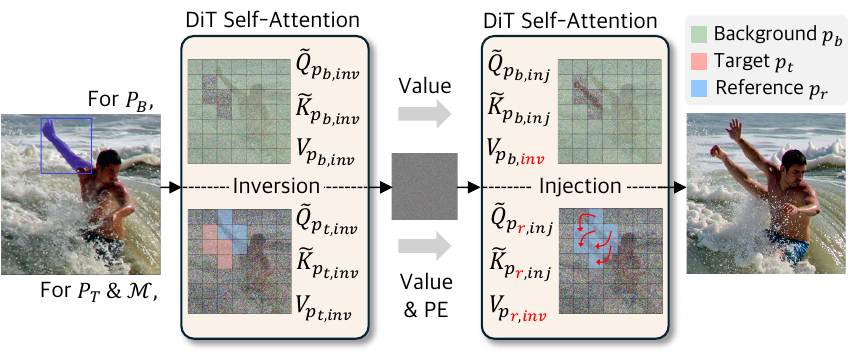}
    \vspace{-0.5cm}
    \caption{Inversion-injection module. In this example, the right arm from the reference patches is added to target patches below it.}
    \label{fig:inversion_injection}
    \vspace*{-0.5cm}
\end{figure}


We propose an \textit{inversion-injection} module, which extends the inversion-restoration paradigm from image editing~\citep{meng2022sdedit, mokady2023null, wang2025rfedit, deng2025fireflow}. Specifically, the inversion-injection module employs the target-reference patch mapping to manipulate the positional information in the self-attention layers of DiT, allowing realistic structural artifact synthesis.

\myparagraph{Notations.}  
Let $X^{(\ell)} \in \mathbb{R}^{N \times d}$ denote the input to a transformer layer $\ell$, where $N$ is the number of image patches and $d$ is the embedding dimensionality. We denote the set of all patch indices as $\mathcal{P}=\{1,\ldots,N\}$. During artifact injection, we accept the target-reference patch mapping $\mathcal{M}=\{(p_t,p_r)\in\mathcal{P}\times\mathcal{P}\}$ provided by the toolbox, where $p_t$ is a target patch to be modified and $p_r$ its reference patch. The set of all target patches is $\mathcal{P}_T = \{p_t \mid (p_t,p_r) \in \mathcal{M}\}$, the set of all reference patches is $\mathcal{P}_R = \{p_r \mid (p_t,p_r) \in \mathcal{M}\}$, and the set of  background patches becomes $\mathcal{P}_B = \mathcal{P} \setminus \mathcal{P}_T$.

\myparagraph{Inversion Stage.} The inversion stage maps the input image into its corresponding noisy latent representation.
Queries, keys, and values for the self-attention layer are
\begin{equation}
Q^{(\ell)} = X^{(\ell)} W_Q^{(\ell)}, \
K^{(\ell)} = X^{(\ell)} W_K^{(\ell)}, \
V^{(\ell)} = X^{(\ell)} W_V^{(\ell)},
\end{equation}
where $W_Q^{(\ell)}, W_K^{(\ell)}, W_V^{(\ell)} \in \mathbb{R}^{d \times d}$ are learnable. We encode positions with rotary embeddings (RoPE)~\citep{su2021rope} such that
\begin{equation}
\tilde{Q}^{(\ell)}_p = \text{RoPE}(Q^{(\ell)}, p), \
\tilde{K}^{(\ell)}_p = \text{RoPE}(K^{(\ell)}, p), \
p \in \mathcal{P}.
\end{equation}
Then, the layer's attention output is
\begin{equation}
\label{eq:self_attn_inversion}
\text{Attn}^{(\ell)}(X^{(\ell)}) =
\text{Softmax}\left(\frac{\tilde{Q}^{(\ell)} \tilde{K}^{(\ell)\top}}{\sqrt{d}}\right) V^{(\ell)}.
\end{equation}
While executing Equation (\ref{eq:self_attn_inversion}), we cache the per-layer value embeddings $V^{(\ell)}_{\text{inv}} \leftarrow V^{(\ell)}$ for use in the injection stage.

\myparagraph{Injection Stage.} 
As in Figure~\ref{fig:inversion_injection}, during the denoising process, the injection stage injects structural artifacts into $\mathcal{P}_T$ by borrowing spatial semantics from $\mathcal{P}_R$, while keeping the original semantics of $\mathcal{P}_B$.

\smallskip
\begin{itemize}[leftmargin=10pt, nosep]
\item
\textbf{Target Region.}  For $p_t\in\mathcal{P}_T$ mapped to $p_r$, we replace the positional embedding\,(PE) and the value embedding of $p_t$ with those of $p_r$: $\tilde{Q}^{(\ell)}_{p_t} = \text{RoPE}(Q^{(\ell)}_{p_t}, p_r)$,
$\tilde{K}^{(\ell)}_{p_t} = \text{RoPE}(K^{(\ell)}_{p_t}, p_r)$,
and $V^{(\ell)}_{p_t} \leftarrow V^{(\ell)}_{p_r,\text{inv}}$.


\item
\textbf{Background Region.} For a non-target patch $p_b \in \mathcal{P}_B$, we keep their original positional information and reuse the $V^{(\ell)}_\text{inv}$ values to maintain the context of the original image: $\tilde{Q}^{(\ell)}_{p_b} = \text{RoPE}(Q^{(\ell)}_{p_b}, p_b)$,
$\tilde{K}^{(\ell)}_{p_b} = \text{RoPE}(K^{(\ell)}_{p_b}, p_b)$,
and $V^{(\ell)}_{p_b} \leftarrow V^{(\ell)}_{p_b,\text{inv}}$.
\end{itemize}

\smallskip
Then, self-attention is performed as in Equation (\ref{eq:self_attn_inversion}).

In general, PE injection controls \emph{where} the model believes denoising is occurring,
while value injection provides \textit{what} semantic content fills that position. Their combination allows for local injections of realistic artifacts, while the background remains consistent with the original image. Although \emph{value injection}~\citep{wang2025rfedit, deng2025fireflow} is a well-established approach in image editing research, our method introduces the novel idea of manipulating their positions during injection. Furthermore, \emph{PE injection} itself has not been used in previous image editing works. Combining PE and value injections offers a particularly effective methodology for generating structural artifacts, as it enables direct manipulation of spatial information during the reconstruction process. 
To prevent shortcut-feature learning (e.g., edge discontinuities), we restrict PE and value injections to early-to-middle layers and disable them during the final denoising steps. We provide attention visualizations and ablation studies (including injection-step ablations) to support these design choices in Appendices~\ref{apdx:visualizations} and \ref{apdx:ablations}. Furthermore, our training-free inversion-injection method is inherently model-agnostic, making it readily generalizable to other emerging DiT-based image generation models.


\subsection{Curation Agent (Figure \ref{fig:method}(c))}
The curation agent is the pipeline's quality assurance and enrichment stage, which refines the synthesis agent's output into training-ready datasets for downstream tasks. Given paired real and artifact-injected images, the curation agent performs data filtering and explanation generation. This paired input enables reliable filtering and explanation by \textit{contrasting} artifact-injected regions with real counterparts.

\subsubsection{Data Filtering} 
\label{subsubsec:data_filtering}
This stage applies one of two filtering methods depending on which tool was used for artifact injection.

\begin{itemize}[leftmargin=10pt,noitemsep]
\item \textbf{LPIPS-Based Filtering.} Distortion artifacts are validated using the LPIPS~\citep{zhang2018lpips} metric, which measures the perceptual difference between two images by comparing their feature representations, producing a metric that aligns much more closely with human judgments. For each cropped original-injected pair, we compute the LPIPS score and retain the pair if it satisfies
$
    \tau_1 \leq 1-d_{\text{LPIPS}}(x_{\text{original}},x_{\text{artifact}})\leq\tau_2,
$ 
with $\tau_2$ filtering out pairs too similar with unidentifiable changes, while $\tau_1$ filters out severely corrupted samples with implausible damage in the region.
\item \textbf{VLM-Based Filtering.} For duplication, omission, and fusion artifacts, we employ a VLM to validate whether the injected change is perceptible and localized within the designated region. The VLM receives the original image with the target region masked out (to provide the global scene context),  the original image cropped to the target region (to show the unaltered content), and the artifact-injected image cropped to the same region (to focus on the modification). From this triplet, the VLM makes a binary judgment, confirming whether a new instance has appeared (duplication), an expected object has become missing (omission), or two objects have been unnaturally merged (fusion). For a detailed prompt template used for VLM-based filtering, see Appendix~\ref{apdx:filtering}.
\end{itemize}

\subsubsection{Explanation Generation}
\label{subsubsec:explanation}
\begin{itemize}[leftmargin=10pt,noitemsep]
\item \textbf{Local Explanation}. For each candidate region, the VLM receives the same triplet used in filtering. We prompt the VLM to synthesize a short local description by guiding the VLM to describe what is different in the artifact region compared to the real counterpart. For the detailed prompt template used, see Appendix~\ref{apdx:explanation}.
\item \textbf{Global Explanation}. After generating local explanations for all artifacts, the curation agent generates a global explanation for the whole image. The VLM accepts the artifact image and a list of artifact-injected bounding boxes and their corresponding local explanations. The VLM is prompted to explain why this artifact injected image is indeed an artifact. For the detailed prompt template used, see Appendix~\ref{apdx:explanation}.
\end{itemize}

\subsection{Data Collection}
With our novel three-stage agentic pipeline, we collect 50K pairs of artifact-injected images and the corresponding original images, along with the metadata consisting of artifact explanation shown in the final output of Figure \ref{fig:method}. Here, we reconstruct the source images using the inversion-restoration method to alleviate pairwise differences originating from diffusion-generated image traits. The source images are composed of four different datasets, COCO~\cite{chen2015mscoco}, Caltech-101~\cite{fei2007learning}, 11K Hands~\cite{afifi201911kHands}, and Celeba HQ~\cite{DBLP:journals/corr/abs-1710-10196}, broadening the distribution from diverse real-world scenes to specific single entity images.
We employ GPT-4o~\cite{openai2024gpt4o} as the VLM that contributes to our agentic flow. In addition, the inversion-injection method in Section \ref{subsubsec:inversion_injection} uses FLUX.1-dev~\cite{FLUX} for the DiT and FireFlow~\cite{deng2025fireflow} for the inversion-injection module.

\section{\benchname{}: Artifact Detection Benchmark}
\label{sec:benchmark}
\begin{table}[t]
\small
\setlength{\tabcolsep}{4pt}
\caption{\textbf{Comparison of artifact benchmark datasets.} If a benchmark reused another dataset as a source, we describe the generative models used in that source. The table with the full citations is provided in Appendix~\ref{apdx:eval_source}.}
\vspace{-0.3cm}
\centering
\resizebox{\columnwidth}{!}{\begin{tabular}{@{}l cc c c c c@{}}
\toprule
\multirow{2}{*}{\textbf{Benchmark}} 
& \multicolumn{2}{c}{\textbf{Source Models}}
& \multirow{2}{*}{\textbf{Sample}} 
& \multirow{2}{*}{\textbf{Bin.}} 
& \multirow{2}{*}{\textbf{Loc.}} 
& \multirow{2}{*}{\textbf{Exp.}} \\
& Oldest & Newest & & & & \\
\midrule
\multirow{2}{*}{RichHF} & \multirow{2}{*}{SD2.1} & Dreamlike & \multirow{2}{*}{955} & \multirow{2}{*}{} & \multirow{2}{*}{\checkmark} & \multirow{2}{*}{} \\
&&Photoreal&&&&\\
\midrule
LOKI & pix2pix & FLUX & 229 & & \checkmark & \checkmark \\
\midrule
SynthScars & Midjourney & DALL$\cdot$E3 & 1K & & \checkmark & \checkmark \\
\midrule
\textbf{\benchname{}} & SD3.5 & Nano-Banana& 1K & \checkmark & \checkmark & \checkmark \\
\bottomrule
\end{tabular}}
\vspace{-0.5cm}
\label{tbl:eval_source}
\end{table}

\newcolumntype{M}{>{\centering\arraybackslash}p{0.08\textwidth}}
\setlength{\tabcolsep}{1pt}

\newcommand{\sotacell}[1]{\cellcolor{blue!15}{#1}}        
\newcommand{\secondsotacell}[1]{\cellcolor{green!15}{#1}} 

\begin{table*}[!t]
\centering
\normalsize
\caption{\textbf{Artifact understanding performance across (a) binary detection, (b) localization, and (c) explanation.}
The benchmarks are listed in the order in which they were published.}\vspace{-0.2cm}

\resizebox{\textwidth}{!}{%
\begin{tabular}{@{}c|
M M |
M M M M M M M M |
M M M M M M @{}}

\toprule
\multirow{3}{*}{Methods}& \multicolumn{2}{c|}{\underline{(a) Binary detection}}
& \multicolumn{8}{c|}{\underline{(b) Localization}}
& \multicolumn{6}{c}{\underline{(c) Explanation}} \\

\multicolumn{1}{c|}{}%
& \multicolumn{2}{c|}{\benchname{}}
& \multicolumn{2}{c}{RichHF}
& \multicolumn{2}{c}{LOKI}
& \multicolumn{2}{c}{SynthScars}
& \multicolumn{2}{c|}{\benchname{}}
& \multicolumn{2}{c}{LOKI}
& \multicolumn{2}{c}{SynthScars}
& \multicolumn{2}{c}{\benchname{}} \\

 &
Acc & F1 &
mIoU & F1 &
mIoU & F1 &
mIoU & F1 &
mIoU & F1 &
ROUGE & CSS &
ROUGE & CSS &
ROUGE & CSS \\
\midrule

PAL &
- & - &
0.079 & 0.028 &
0.021 & 0.037 &
0.035 & 0.053 &
0.040 & 0.066 &
- & - &
- & - &
- & - \\

DiffDoctor &
- & - &
0.077 & 0.139 &
\textbf{0.175} & \textbf{0.274} &
0.083 & 0.136 &
0.081 & 0.137 &
- & - &
- & - &
- & - \\

LEGION &
- & - &
0.067 & 0.112 &
0.100 & 0.158 &
0.106$^{\dagger}$ & 0.152$^{\dagger}$ &
0.062 & 0.099 &
0.133 & 0.314 &
\textbf{0.247}$^{\dagger}$ & \textbf{0.589}$^{\dagger}$ &
0.143 & 0.332 \\

\midrule
GPT-4o &
0.619 & 0.601 &
0.086 & 0.040 &
0.037 & 0.056 &
0.032 & 0.052 &
0.049 & 0.084 &
0.107 & 0.266 &
0.125 & 0.404 &
0.143 & 0.433 \\

Gemini-2.5-Pro &
0.582 & 0.575 &
0.061 & 0.091 &
0.109 & 0.169 &
0.064 & 0.101 &
0.095 & 0.147 &
0.097 & 0.358 &
0.103 & 0.474 &
0.159 & 0.420 \\

GPT-5 &
0.599 & 0.577 &
\textbf{0.126} & 0.146 &
0.089 & 0.141 &
0.117 & 0.185 &
0.061 & 0.099 &
0.121 & 0.382 &
0.120 & 0.461 &
0.145 & 0.434 \\

\midrule
Qwen2.5-VL-7B &
0.501 & 0.336 &
0.075 & 0.028 &
0.052 & 0.068 &
0.013 & 0.018 &
0.010 & 0.014 &
0.106 & 0.267 &
0.115 & 0.362 &
0.117 & 0.263 \\

\rowcolor{gray!15}
+ \tblalgname{} &
\underline{0.627} & \textbf{0.627} &
\underline{0.119} & \textbf{0.198} &
\underline{0.129} & \underline{0.198} &
\underline{0.137} & \underline{0.214} &
\underline{0.111} & \underline{0.168} &
\textbf{0.169} & \textbf{0.454} &
\underline{0.196} & \underline{0.578} &
\textbf{0.233} & \textbf{0.643} \\

InternVL3.5-8B &
0.498 & 0.357 &
0.013 & 0.022 &
0.015 & 0.025 &
0.019 & 0.033 &
0.010 & 0.015 &
0.081 & 0.189 &
0.050 & 0.180 &
0.126 & 0.256 \\

\rowcolor{gray!15}
+ \tblalgname{} &
\textbf{0.630} & \underline{0.620} &
0.100 & \underline{0.170} &
0.126 & 0.196 &
\textbf{0.140} & \textbf{0.217} &
\textbf{0.119} & \textbf{0.176} &
\underline{0.137} & \underline{0.401} &
0.179 & 0.513 &
\underline{0.226} & \underline{0.625} \\

\bottomrule
\end{tabular}
\label{tbl:vlm_all}
}

{\raggedright\scriptsize
$^{\dagger}$LEGION was trained on the SynthScars training dataset split.
\par}

\vspace{-0.4cm}
\end{table*}

Existing artifact detection benchmarks, such as RichHF~\citep{kirstain2023pickapicopendatasetuser}, LOKI~\citep{ye2024loki}, and SynthScars~\citep{kang2025legion} were built primarily using earlier diffusion models, including Stable Diffusion 1 or 2 and Midjourney. These datasets, though valuable at the time, no longer capture the characteristics of modern artifacts produced by current diffusion and multimodal transformers. As a result, the evaluation on these benchmarks may not accurately reflect the artifact-handling capabilities of today's models.

To address this issue, we introduce \textbf{\benchname{}}, a new benchmark that reflects the current state of artifact phenomena in recent generative models. As summarized in Table~\ref{tbl:eval_source}, previous benchmarks were limited either in sample diversity, recency of generative sources, or task coverage. Our benchmark is designed to overcome these limitations by including data from recent models and providing comprehensive annotations for multiple artifact-related tasks.

We construct \textbf{\benchname{}} with 1K images generated by five state-of-the-art diffusion models, Stable Diffusion3.5~\citep{esser2024sd3}, FLUX-schnell/dev~\citep{FLUX}, Qwen-Image~\citep{wu2025qwenimage}, and Nano-Banana~\citep{google2025nanobanana}, with the prompts sampled from three datasets, MS-COCO~\citep{chen2015mscoco}, PartiPrompts~\citep{yu2022parti}, and FuseCap~\citep{rotstein2024fusecap}. For annotation, we involve 12 human annotators and label each image with:
(1) a binary indicator denoting the presence or absence of artifacts,
(2) bounding boxes for all artifact regions, and
(3) concise descriptions of the observed abnormalities.
The dataset is balanced with an equal ratio of artifact-free and artifact-containing samples. Further details on the construction and annotation pipeline are provided in Appendix~\ref{apdx:bench_gen}.
\section{Experiments}
\label{sec:exp}

\subsection{Understanding Artifacts with VLMs}
\label{subsec:vlm_eval}

We assess the efficacy of \algname{} by training VLMs with visual question-answering\,(VQA) samples generated from \algname{} and measuring their performance on artifact-aware benchmarks, including \benchname{}.

\subsubsection{Setup} We consider \textit{three} tasks, artifact detection, localization, and explanation. For evaluation metrics, we use accuracy and F1 score for detection, mIoU and F1 score for localization, and ROUGE and CSS for explanation. More details of the evaluation protocol are provided in Appendix~\ref{apdx:eval_process}.
For baselines, we use three artifact segmentation algorithms, PAL~\citep{zhang2023pal}, DiffDoctor~\citep{wang2025diffdoctor}, and LEGION~\citep{kang2025legion}; and three proprietary VLMs, GPT-4o~\citep{openai2024gpt4o}, Gemini-2.5-Pro~\citep{deepmind2025gemini2.5pro}, and GPT-5~\citep{openai2025gpt5}.
PAL and DiffDoctor are evaluated only on the localization task as they output only segmentations, LEGION is evaluated on the localization and explanation tasks, and the three VLMs are evaluated on all three tasks. We fine-tune Qwen2.5-VL-7B~\citep{bai2025qwen2.5vl} and InternVL3.5-8B~\citep{zhu2025internvl3} on a 100K training set generated by \algname{}. The detailed procedure for  VQA generation is provided in Appendix~\ref{apdx:vlm_training}.

\subsubsection{Main Results}

Overall, fine-tuning open-source VLMs with synthetic data generated by \algname{} consistently enhances their ability to detect, explain, and localize visual artifacts. Across all three tasks, \algname{}-trained models not only outperform their vanilla counterparts but also match or exceed the performance of proprietary systems such as GPT-5 and Gemini-2.5-Pro. These gains highlight the quality and scalability of \algname{}-generated supervision, demonstrating that artifact synthesis through agentic data generation provides rich, transferable signals for both spatial grounding and semantic reasoning.

\myparagraph{Artifact Binary Detection.}
Table~\ref{tbl:vlm_all}(a) shows that VLMs trained with \algname{} achieve clearly superior artifact detection performance.
Specifically, \algname{} improves the accuracy of InternVL3.5-8B by 26.5\%.
At the same time, the overall low accuracy on \benchname{} highlights the difficulty and importance of understanding visual artifacts in AI-generated images, especially because modern diffusion models increasingly exhibit subtle, structured failures that current VLMs easily miss.

\myparagraph{Artifact Localization.}
Table~\ref{tbl:vlm_all}(b) shows that \algname{} consistently enhances the spatial grounding capability of open-source VLMs. Although DiffDoctor exhibits high accuracy in LOKI, it fails to generalize to more recent benchmarks, including \benchname{}. This outcome means that \benchname{} further extends the artifact detection field one step deeper, by capturing artifacts that prior artifact-detection models struggle to identify.

\myparagraph{Artifact Explanation.}
Table~\ref{tbl:vlm_all}(c) shows that training with the \algname{}'s generated dataset greatly strengthens the reasoning and description capabilities of VLMs. The fine-tuned models exhibit clear improvements in the ROUGE and CSS metrics across all benchmarks.

\begin{figure}[t]
    \centering
    \vspace{-0.6cm}
    \includegraphics[width=\linewidth]{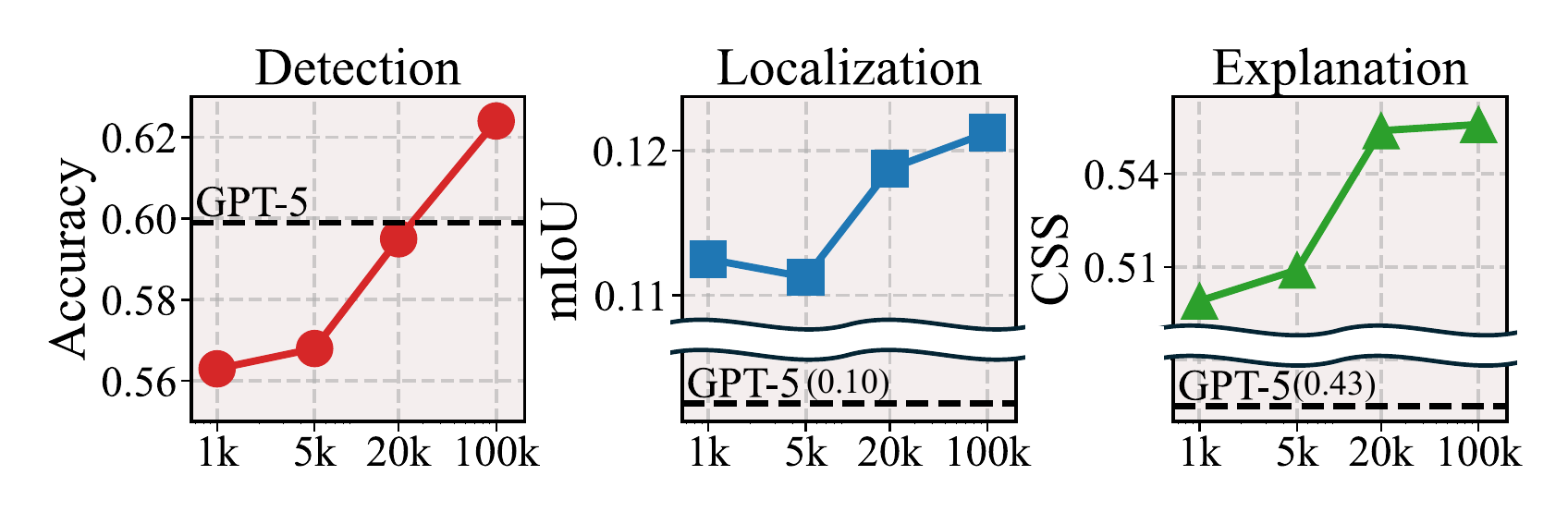}
    \vspace{-1.0cm}
    \caption{\textbf{Scaling effect} of data generated by \algname{} with Qwen2.5-VL-7B. We average the results of all the benchmarks.}
    \label{fig:scaling_law}
    \vspace{-0.3cm}
\end{figure}
\begin{table}[t]
\small
\setlength{\tabcolsep}{2pt}
\caption{ArtiBench result of the VLM trained on a 1K dataset.}
\label{tbl:compact_1k}
\vspace{-0.35cm}
\centering
\renewcommand{\arraystretch}{0.95}
\resizebox{\columnwidth}{!}{%
\scriptsize{
\begin{tabular}{@{}c|>{\centering\arraybackslash}p{0.11\columnwidth}>{\centering\arraybackslash}p{0.11\columnwidth}|>{\centering\arraybackslash}p{0.11\columnwidth}>{\centering\arraybackslash}p{0.11\columnwidth}|>{\centering\arraybackslash}p{0.11\columnwidth}>{\centering\arraybackslash}p{0.11\columnwidth}@{}}
\toprule
\multirow{2}{*}{Methods} &
\multicolumn{2}{c|}{\underline{(a) Binary detection}} &
\multicolumn{2}{c|}{\underline{(b) Localization}} &
\multicolumn{2}{c}{\underline{(c) Explanation}} \\
& Acc & F1 & mIoU & F1 & ROUGE & CSS \\
\midrule
SynthScars & \textbf{0.555} & 0.548 & \textbf{0.094} & \textbf{0.147} & 0.156 & 0.521 \\
\cellcolor{gray!15}\textbf{ArtiAgent} & \cellcolor{gray!15}\textbf{0.555} & \cellcolor{gray!15}\textbf{0.550} & \cellcolor{gray!15}0.074 & \cellcolor{gray!15}0.121 & \cellcolor{gray!15}\textbf{0.222} & \cellcolor{gray!15}\textbf{0.606} \\
\bottomrule
\end{tabular}%
}}
\vspace{-0.6cm}
\end{table}


\subsubsection{Data Scaling Effect of \algname{}}
Figure~\ref{fig:scaling_law} shows how model performance grows as we increase the size of the training data with \algname{}. In all three tasks, we observe a clear upward trend, which means that more synthesized data consistently lead to better artifact understanding. Notably, for localization and explanation, the performance with subsets as small as 1K samples already surpasses that of GPT-5, showing that \algname{} provides highly sample-efficient supervision. In contrast, the binary detection task continues to improve up to the 100K scale, suggesting that detection benefits from larger and more diverse artifacts. These results highlight the rich supervision and the strong scaling potential of \algname{}.

\subsubsection{Comparison with Human-Annotated Supervision}
To compare the quality of human annotation and ArtiAgent-generated annotations, we train Qwen2.5-VL-7B with 500 SynthScars samples and 500 clean samples from ArtiAgent and compare it with a counterpart trained on 1K ArtiAgent samples. Table~\ref{tbl:compact_1k} shows that ArtiAgent's synthetic supervision remains competitive with human annotation: SynthScars is slightly better in localization, while ArtiAgent matches in detection and performs better in explanation. We attribute the localization gap mainly to the patch-level granularity of ArtiAgent labels. Overall, these results indicate that ArtiAgent is a cost-effective and scalable alternative to human annotation while maintaining comparable supervision quality.

\subsection{Mitigating Artifacts in Diffusion Models}
\subsubsection{Reward-Guided Artifact-Free Generation}
\label{subsubsec:reward_guided_gen}

A key strength of \algname{} is its pairwise data design: for every instance, it provides two tightly matched images with the same content, one clean and one with artifact. This structure gives extremely rich supervision for learning artifact preferences. Using these pairs, we adopt the reward-guided test-time scaling framework~\cite{ma2025ttdiffusion} that steers diffusion models toward producing artifact-free images.

\myparagraph{Setup.}
We use CLIP as the backbone of the Bradley-Terry~\cite{btmodel} reward model. The model learns to assign a higher score to the real image over the artifact image. With this artifact-aware reward model, we apply test-time scaling to FLUX-schnell. We run six search rounds, each for 100 prompts sampled from MS-COCO and measure how much the reward increases as the search progresses. The detailed training schema and the test-time scaling procedure are provided in the Appendix~\ref{apdx:reward_guided_gen}.

\begin{figure}[t]
    \centering
    \includegraphics[width=\linewidth]{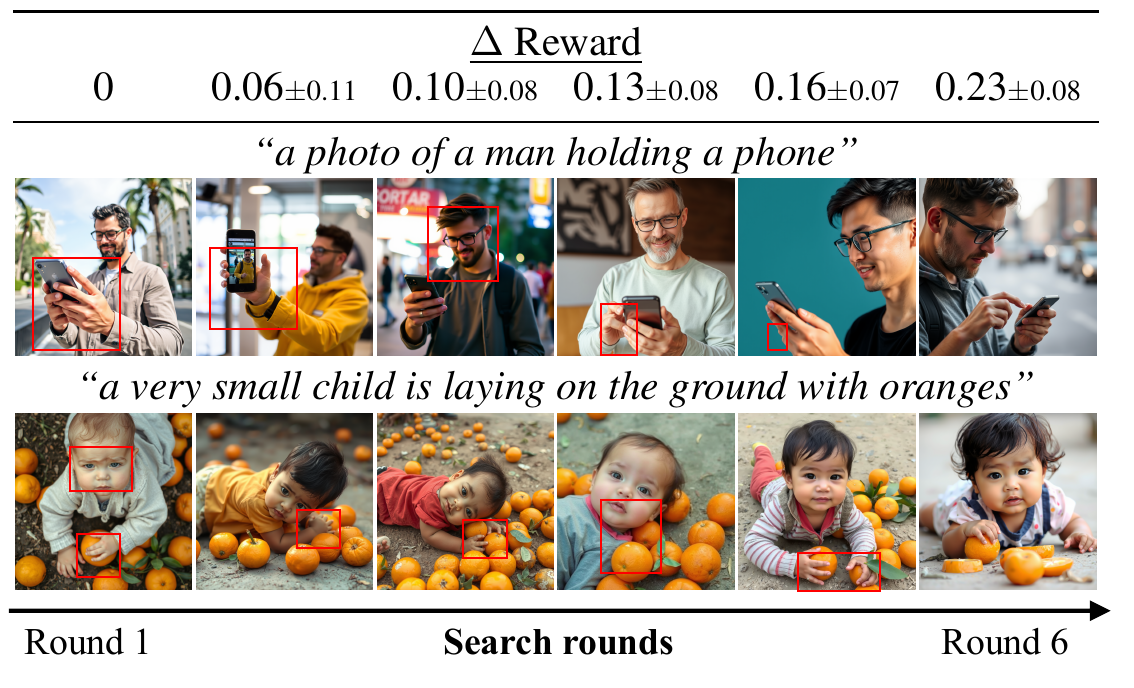}
    \vspace{-0.7cm}
    \caption{\textbf{Reward-guided generation.} \algname{} can train a reward model that guides diffusion to generate artifact-free images.}
    \label{fig:reward_guided_gen}
    \vspace{-0.5cm}
\end{figure}

\myparagraph{Results.} 
As can be seen in Figure~\ref{fig:reward_guided_gen}, the reward steadily improves throughout the search rounds, indicating that the diffusion model continues to generate images with fewer artifacts. Qualitatively, the examples in Figure~\ref{fig:reward_guided_gen} show clearer structures and reduced artifact patterns in later rounds, demonstrating that the reward model has successfully learned the real-artifact preference and enables guidance towards artifact-free images.
\smallskip

\subsubsection{VLM-Guided Artifact Correction}
\label{subsubsec:artifact_correction}

Since our artifact-trained VLM can reliably detect and localize artifacts, we employ it to guide an image inpainting model to correct artifact regions in AI-generated images.

\myparagraph{Setup.}
We use Qwen2.5-VL-7B trained with \algname{} to determine whether an artifact is present in a given image and to localize the artifact region. This region is then processed by the FLUX inpainting pipeline~\cite{fluxinpaint}, which synthesizes a corrected version of the localized area. Next, the corrected image is re-evaluated by the VLM to verify whether the artifact in the region has been fully resolved in the specified region. If the VLM continues to detect an artifact, the inpainting procedure is repeated. This iterative loop continues until the VLM confirms the absence of artifacts. The detailed procedure is provided in the Appendix~\ref{apdx:image_correction}.

\begin{figure}[t]
    \centering
    \includegraphics[width=\linewidth]{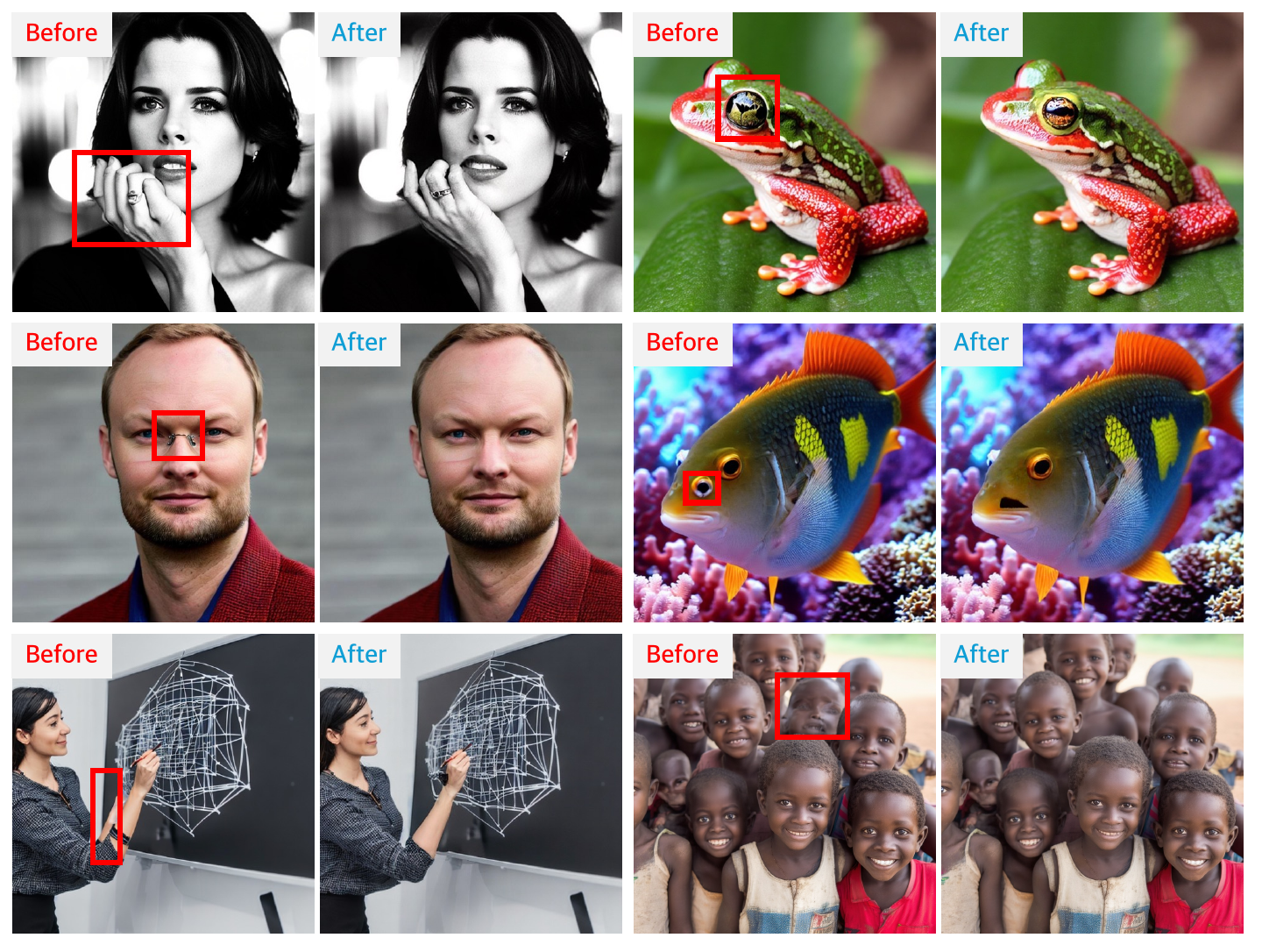}
    \vspace{-0.5cm}
    \caption{\textbf{Image correction.} The \algname{}-trained VLM can effectively guide image inpainting models to correct artifact regions.}
    \label{fig:correction}
    \vspace{-0.1cm}
\end{figure}

\myparagraph{Results.}
Figure~\ref{fig:correction} presents qualitative results of our image correction pipeline. The VLM accurately identifies the artifact region and the inpainting model corrects the region with natural and structurally consistent content. These results demonstrate that the proposed pipeline can reliably locate and correct artifacts, depicting the usefulness of our artifact understanding VLM.

\section{Conclusion}
\label{sec:con}
In this work, we introduce \algname{}, a scalable agentic framework that automatically synthesizes visual artifacts through positional embedding manipulation in diffusion transformers. By integrating perception, synthesis, and curation agents, our pipeline generates large-scale, richly annotated artifact datasets without human supervision. Experiments showed that VLMs fine-tuned on the \algname{} datasets achieved substantial gains in artifact detection, localization, and explanation, and that the resulting models can guide diffusion sampling toward artifact-free generations and perform automated artifact correction. Together, these results demonstrate that agentic data synthesis provides an effective and general pathway to perceiving and mitigating visual artifacts in modern generative models.



\newpage
\section*{Acknowledgements}
This work was supported by Institute of Information \& Communications Technology Planning \& Evaluation\,(IITP) grant funded by the Korea government\,(MSIT) (No.\ RS-2020-II200862, DB4DL: High-Usability and Performance In-Memory Distributed DBMS for Deep Learning, 50\% and No.\ RS-2025-25442149, LG AI STAR Talent Development Program for Leading Large-Scale Generative AI Models in the Physical AI Domain, 50\%).

{
    \small
    \bibliographystyle{ieeenat_fullname}
    \bibliography{reference}
}
\clearpage
\newpage

\appendix
\setcounter{page}{1}
\maketitlesupplementary

\section{\algname{} Pipeline Details (\S~\ref{sec:method})}
\label{apdx:pipeline}

\subsection{Implementation Details}
\label{apdx:impl_details}
\myparagraph{Injection Stage.} The details of the injection process in Section \ref{subsubsec:inversion_injection} were designed to introduce natural and coherent structural artifacts to the image. We accept and build on the coarse-to-fine manner of denoising processes, adapting the understanding that earlier denoising time steps contribute to overall structural context, while later steps focus on regions of fine-grained details~\cite{agarwal2025towards, park2024explaining, park2023understanding}.
\begin{itemize}
    \item \textbf{PE Injection.} Among the 25 denoising steps, we disable PE injection for selected final steps and apply PE injection only in the earlier steps. Duplication, distortion, and fusion artifacts disable five final steps, focusing on introducing coarse context of structural modifications while maintaining natural connection with the original scene. In contrast, the omission artifact is generated by disabling PE injection only in the final timestep, since it requires stronger perturbation on  positional context to remove an existing feature from the image and distinguish the newly injected background from its original entity.
    \item \textbf{Value Injection.} Value injection is performed only for the first 15 denoising steps. Reducing the injection steps for value injection ensures artifact injection while maintaining image quality. For the architecture consisting of sequential double stream blocks and single stream blocks of FLUX.1-dev, only the deeper single stream blocks (20-38) carry out the value injection process. 
\end{itemize}
\myparagraph{Filtering}. 
To ensure alignment with human visual perception, the distortion-type artifact filtering thresholds, $\tau_1$ and $\tau_2$, were heuristically established at 0.5 and 0.9, respectively. For each artifact region cropped from the original image and the artifact-injected image, a LPIPS distance was measured to ensure certain quality among the artifacts introduced. While malformed regions that are not acceptable as plausible structural artifacts are filtered out by a high LPIPS distance over 0.5, unsuccessful artifact injections with similar features are discarded by the low LPIPS distance, indicating high similarity.

\subsection{Prompt Templates for \algname{}}
\subsubsection{Entity-Subentity Vocabulary (\S~\ref{subsubsec:entity_subentity_vocab})}
\label{apdx:ent_subent}
We employ the capability of GPT-4o~\citep{openai2024gpt4o} to identify and recognize relationships between entities in images. To ensure that the extracted entity-subentity sets are clearly segmentable and valid for generating plausible artifacts, we set strict rules and guidance as in Figure \ref{fig:vocabs_helper} and instruct the VLM model to respond with qualified sets of vocabulary in an explicit format. Figure \ref{fig:vocabs} shows the precise prompt used to generate the entity-subentity vocabulary sets and an example response from the perception agent. 

\subsubsection{Data Filtering (\S~\ref{subsubsec:data_filtering})}
\label{apdx:filtering}
The curation agent uses a set of deliberately described artifact types referring to the injection methods and detailed instructions to detect or explain the artifacts, shown in Figure \ref{fig:artifact_types}. Utilizing the descriptions and criteria according to the type of artifact injected, we query the VLM model with the entity name and a triplet of images consisting of (1) the original image with the target part masked out, (2) the original image with only the target part cropped, and (3) the generated image with only the target part cropped. 

Although modern out-of-the-box VLMs demonstrate limited understanding of structural artifacts, providing the models with the rich image context through the image triplet and detailed descriptions of \textit{the type of artifact} enhances their reliability for this task. The prompt template is elaborated in Figure \ref{fig:detection}.

\subsubsection{Explanation Generation (\S~\ref{subsubsec:explanation})}
\label{apdx:explanation}
Similarly to Appendix~\ref{apdx:filtering}, we employ artifact type description in Figure~\ref{fig:artifact_types} to generate rich explanations. The detailed prompt and an example of the local and global explanation generation process is shown in Figure~\ref{fig:expl_regional} and Figure~\ref{fig:expl_global}. By providing multiple images that contain both the global and local view, we maximize VLM capacity of understanding structural artifacts for reliable quality in explanations. Moreover, we employ BLIP2~\cite{li2023blip} to generate concise caption describing the real image.

\subsection{Synthesis Agent Tools (\S~\ref{subsubsec:toolbox})}
\label{apdx:gen_tools}

The artifact injection tools in the synthesis agent produce target–reference patch mappings that are subsequently consumed by the inversion–injection module in Section~\ref{subsubsec:inversion_injection}. Each tool follows a common interface but implements a different geometric prior tailored to a specific artifact type (duplication, omission, distortion, and fusion).

\myparagraph{Notation.}
Let the image be discretized into a patch grid of size $(h_p, w_p)$, and let $\mathcal{P}_{\text{all}}$ be the set of all patch coordinates on this grid. For a given tool call, the tool outputs a target–reference mapping
\[
\mathcal{M} = \{(p_t, p_r)\},
\]
where $p_t \in \mathcal{P}_{\text{all}}$ denotes a \emph{target} patch to be modified, and $p_r \in \mathcal{P}_{\text{all}}$ is its \emph{reference} patch whose semantics will be injected at $p_t$ during the diffusion inversion–injection stage. We freely switch between linear indices and $(y,x)$ grid coordinates using simple index–coordinate conversion routines.

For tools that operate on entity- or subentity-specific regions, we additionally assume access to patch sets such as the same-entity foreground $\mathcal{P}_{\text{ent}}$ and same-subentity foreground $\mathcal{P}_{\text{sub}}$. Concretely, $\mathcal{P}_{\text{ent}}$ contains all patches that belong to a single object instance (e.g., one person or one dog), while $\mathcal{P}_{\text{sub}} \subseteq \mathcal{P}_{\text{all}}$ marks patches of other instances of the same semantic part (e.g., hands of other people, paws of other dogs) that we want to avoid colliding with. These sets are provided by the synthesis agent’s perception stage (e.g., derived from instance/part segmentation) and are treated as fixed inputs when running the tools.

Whenever we say that we \emph{clip to the valid patch grid}, we mean that any candidate coordinate $(i,j)$ whose row or column index falls outside the range $0 \le i \langle   h_p$ or $0 \le j \langle   w_p$ is either discarded or projected back into the rectangular domain $[0, h_p - 1] \times [0, w_p - 1]$ by truncating it to the nearest boundary index, so that all patches used by the tools lie on valid positions of the patch grid.

\myparagraph{Add Tool (duplication).}
The Add Tool (Algorithm~\ref{alg:add}) realizes duplication-type artifacts by creating an extra, plausibly placed copy of a subentity (e.g., an extra hand or paw adjacent to the original one). Given a set of reference patches $\mathcal{P}_R$ for the original subentity, the tool first computes the subentity centroid $(c_i, c_j)$ in patch space and constructs a perimeter band $\mathcal{P}_{\text{ring}}$ of candidate locations around this centroid with Manhattan distance in $[1, \alpha]$. For each candidate $(i,j) \in \mathcal{P}_{\text{ring}}$, it evaluates the score
\[
S(i,j) = \bigl(3 - r_{\text{self}} - r_{\text{ent}} - r_{\text{sub}}\bigr)\, g_{\text{dist}},
\]
where $r_{\text{self}}$, $r_{\text{ent}}$, and $r_{\text{sub}}$ measure the overlap ratio of the shifted subentity with (i) the original subentity region itself, (ii) other foreground patches of the same entity, and (iii) foreground patches belonging to other instances of the same subentity, respectively, and $g_{\text{dist}}$ is a distance-based decay term that penalizes large offsets. Intuitively, this prefers candidate locations that (i) stay close to the source entity, (ii) minimally overlap the original instance, and (iii) avoid collisions with other same-subentity regions. The tool then selects the best-scoring perimeter patch $(i^\star,j^\star)$, computes the corresponding offset $(\Delta_i^\star,\Delta_j^\star)$, and builds the mapping
\[
\mathcal{M} = \{((r_i+\Delta_i^\star, r_j+\Delta_j^\star), (r_i,r_j)) : (r_i,r_j)\in\mathcal{P}_R\},
\]
which duplicates the entire subentity at the chosen location.

\myparagraph{Remove Tool (omission).}
The Remove Tool (Algorithm~\ref{alg:remove}) implements omission-type artifacts by erasing a subentity and filling the region with nearby background context. The target set $\mathcal{P}_T$ is the set of patch coordinates belonging to the subentity being removed. The tool constructs a local neighborhood
\[
\mathcal{P}_{\text{nbr}} = \{\,p \in \mathcal{P}_{\text{all}} : \|p - p_t\|_1 \le R, p_t \in \mathcal{P}_T,\; p \notin \mathcal{P}_T\,\},
\]
discarding any coordinates outside the valid patch grid.

From this neighborhood it derives:
\[
\mathcal{P}_{\text{nbr-no-sub}} = \mathcal{P}_{\text{nbr}} \setminus \mathcal{P}_{\text{sub}}, 
\qquad
\mathcal{P}_{\text{nbr-non-ent}} = \mathcal{P}_{\text{nbr}} \setminus \mathcal{P}_{\text{ent}}.
\]

If sufficiently many true background patches exist, i.e.
\[
|\mathcal{P}_{\text{nbr-non-ent}}| \rangle   \tfrac{1}{2}|\mathcal{P}_{\text{nbr-no-sub}}|,
\]
the tool prioritizes them as the reference pool 
\(\mathcal{P}_R^{\text{pool}} = \mathcal{P}_{\text{nbr-non-ent}}\);  
otherwise it uses 
\(\mathcal{P}_R^{\text{pool}} = \mathcal{P}_{\text{nbr-no-sub}}\),  
which avoids collisions with other same-subentity patches but may include same-entity foreground.

Finally, each target patch $p_t \in \mathcal{P}_T$ selects the nearest reference patch under $L_1$ distance:
\[
p_r = \arg\min_{p \in \mathcal{P}_R^{\text{pool}}} \|p_t - p\|_1,
\]
and the Remove Tool outputs the mapping $\mathcal{M} = \{(p_t, p_r)\}$.

\myparagraph{Distort Tool (distortion).}
The Distort Tool (Algorithm~\ref{alg:distort}) performs structural perturbations within a subentity while keeping its global placement intact. Here, the target and reference sets are drawn from the same foreground region: $\mathcal{P}_T$ indexes the original subentity patches, and $\mathcal{P}_R$ is obtained by applying a distortion kernel. The tool supports three kernel types:
\begin{itemize}
    \item \textbf{Shuffle kernel.} The simplest kernel copies $\mathcal{P}_T$ into $\mathcal{P}_R$ and applies a random permutation. Each target patch is thus reassigned to a randomly chosen patch of the same subentity, breaking local structure while preserving appearance statistics.
    \item \textbf{Gaussian jitter kernel.} For each $(p_y, p_x) \in \mathcal{P}_T$, the kernel repeatedly samples a discrete offset from a Gaussian distribution in patch space, $(\delta_y,\delta_x) \sim \mathcal{N}(0,\sigma^2 I)$. If a sampled location falls within the same-entity foreground $\mathcal{P}_{\text{ent}}$, it is accepted as the reference; otherwise, the kernel resamples up to a fixed budget and falls back to the nearest foreground (or self) patch if necessary. This yields small, local displacements that bend the entity’s internal geometry.
    \item \textbf{Strip kernel.} This kernel first computes the bounding box of $\mathcal{P}_T$ and chooses a dominant direction (vertical or horizontal) by aspect ratio. It then partitions the subentity into $S$ strips along that direction and, within each strip, imposes an ordering of patches. Each strip is circularly shifted by an integer offset $\Delta_s$ (with alternating signs and magnitudes), and the shifted positions define the references. This produces band-like shearing or sliding artifacts within the entity.
\end{itemize}
After applying the chosen kernel to form $\mathcal{P}_R$, the tool returns the one-to-one mapping $\mathcal{M} = \{(\mathcal{P}_T[i], \mathcal{P}_R[i])\}$, which causes the inversion–injection module to reconstruct a structurally distorted yet context-consistent object.

\myparagraph{Fuse Tool (fusion).}
The Fuse Tool (Algorithms~\ref{alg:fuse-main}–\ref{alg:fuse-helpers}) introduces fusion artifacts along the interface of two overlapping entities with patch sets $\mathcal{P}_A$ and $\mathcal{P}_B$. It first identifies the overlap region
\[
\mathcal{P}_{\text{overlap}} = \mathcal{P}_A \cap \mathcal{P}_B
\]
and the union foreground $\mathcal{P}_{\text{fg}} = \mathcal{P}_A \cup \mathcal{P}_B$, as well as the non-overlapping parts $\mathcal{P}_{A\setminus B} = \mathcal{P}_A \setminus \mathcal{P}_{\text{overlap}}$ and $\mathcal{P}_{B\setminus A} = \mathcal{P}_B \setminus \mathcal{P}_{\text{overlap}}$. If $\mathcal{P}_{\text{overlap}} = \emptyset$, no fusion is applied. Given the overlap, the tool constructs a thin fusion band $\mathcal{P}_T$ around $\mathcal{P}_{\text{overlap}}$ by dilating each overlap patch within an $L_1$ radius $R$ and intersecting with $\mathcal{P}_{\text{fg}}$.

To avoid treating the band as a single global region, the tool selects up to $K$ seeds on $\mathcal{P}_T$ via farthest-point sampling and assigns each band patch to its nearest seed in $L_1$ distance, forming local regions $\{\mathcal{R}_s\}$. For each region, it then chooses an \emph{opposite-side pool} $\mathcal{P}_{\text{opp}}$: if the region’s seed is closer (in $L_1$) to $\mathcal{P}_{A\setminus B}$, the pool is $\mathcal{P}_{B\setminus A}$, and vice versa; if distances tie or a side is empty, the pool defaults to $\mathcal{P}_{\text{fg}} \setminus \mathcal{P}_T$. Over a discrete set of small integer offsets
\[
\Omega = \{(\Delta_i,\Delta_j) : 1 \le |\Delta_i| + |\Delta_j| \le R_{\text{off}}\},
\]
the tool then selects the offset $\Delta^\star$ which, when applied to patches in the region, maps the largest number of them onto valid, non-band patches in $\mathcal{P}_{\text{opp}}$. Each band patch is finally paired either with its offset-shifted opposite patch (if valid) or with the nearest patch in $\mathcal{P}_{\text{opp}}$ in $L_1$ distance. The resulting mapping $\mathcal{M}$ injects appearance from one entity into the boundary band of the other, producing visually plausible yet structurally implausible fusion along their interface. In practice, we optionally add a subset of reversed pairs $(p_r, p_t)$ while ensuring that each target is unique, which yields more symmetric blending as visualized in Figure~\ref{fig:tool_viz}.

\subsection{Qualitative Visualizations}
\label{apdx:visualizations}

\begin{figure}[h]
    \centering
    \includegraphics[width=\linewidth]{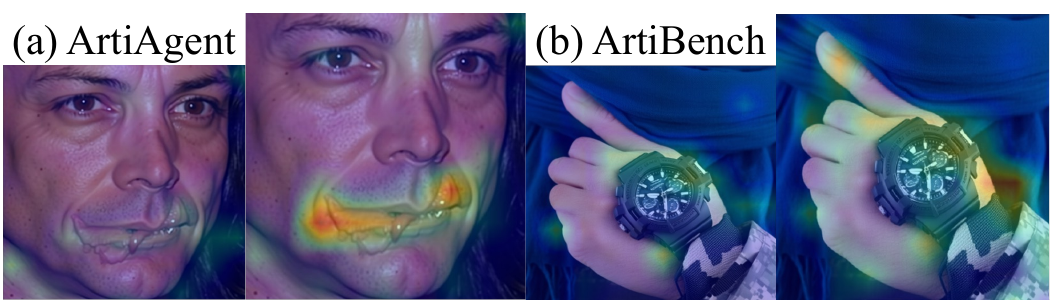}
    \caption{\textbf{Attention Visualization.} We compare the attention maps of InternVL3.5-8B before (base) and after (fine-tuned) training on \algname{}. The fine-tuned model reliably focuses on genuine artifact features across both synthesized images and real-world images.}
    \label{fig:attn_map}
\end{figure}

To validate the coverage and generalization of our synthesized artifacts to natural ones, we analyze attention maps for InternVL3.5-8B before and after fine-tuning with ArtiAgent data. As shown in Figure~\ref{fig:attn_map}, the fine-tuned model precisely attends to artifact regions in both synthetic (ArtiAgent) and real (ArtiBench) images. This comparison confirms that our ArtiAgent-trained VLM successfully learns general artifact features rather than relying on shortcut features such as edge discontinuities, thereby demonstrating strong distributional alignment.

\begin{figure*}[t]
    \centering
    \includegraphics[width=\textwidth]{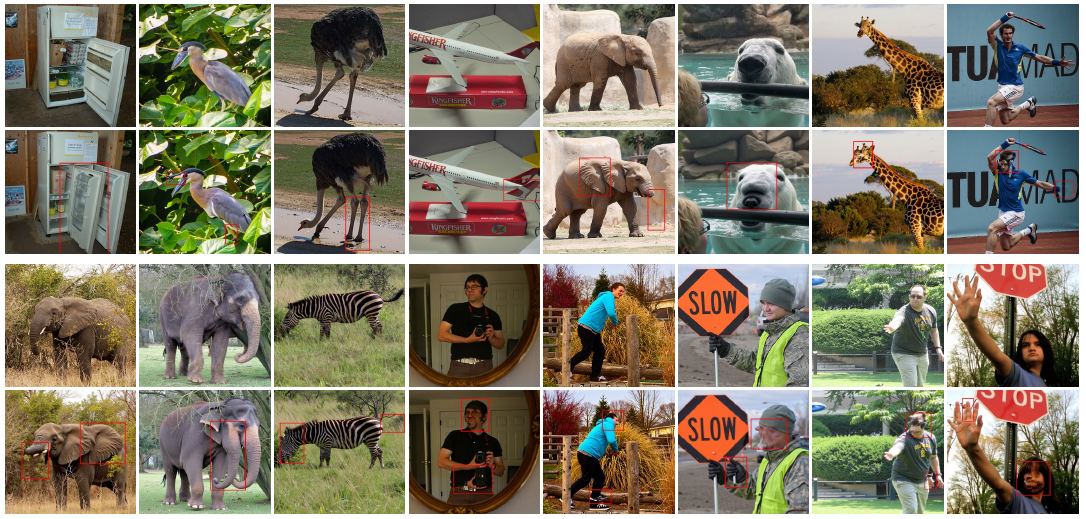}
    \vspace{-0.7cm}
    \caption{Visualizations of artifacts injected with \algname{}. The \textbf{first and third rows} show the original images, whereas the \textbf{second and fourth rows} show the output images with artifacts injected.}
    \label{fig:artiagent_viz}
    \vspace*{-0.3cm}
\end{figure*}

\begin{figure*}
\centering
\includegraphics[width=.98\linewidth]{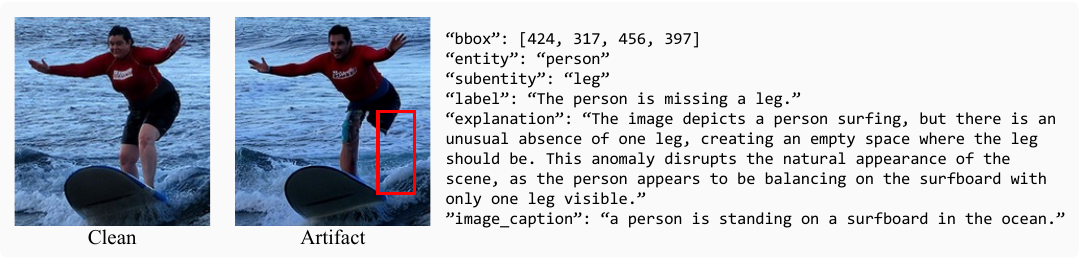}
\vspace{-0.4cm}
\caption{An instance of \algname{} with the annotation.}
\label{fig:final_results}
\end{figure*}

\begin{figure*}[t]
    \centering
    \includegraphics[width=.98\linewidth]{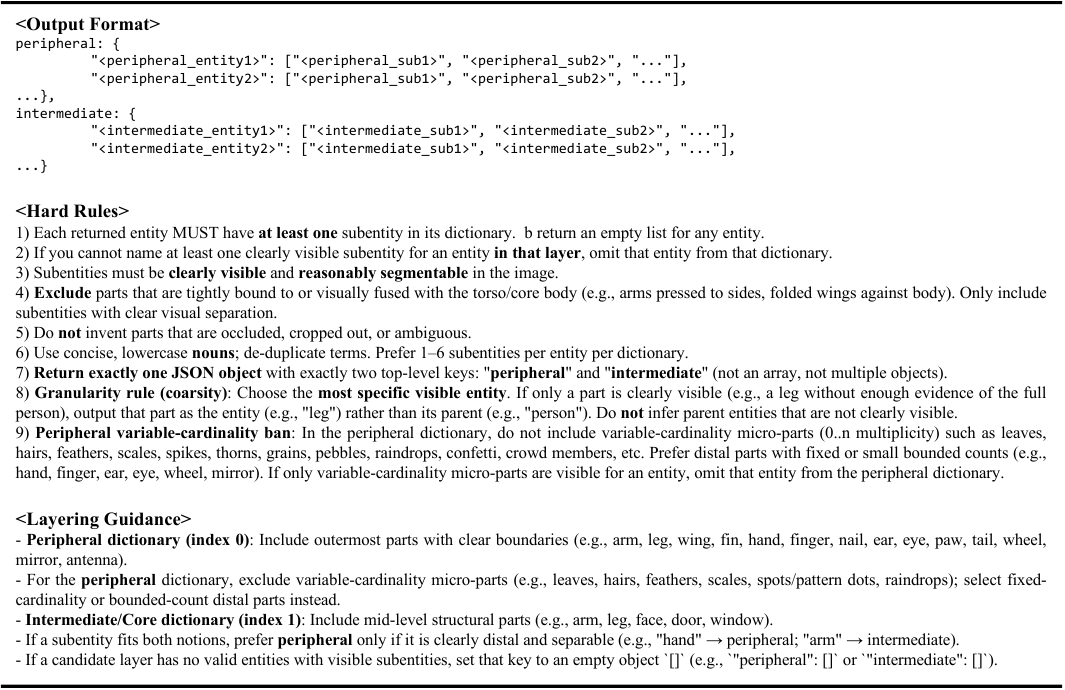}
    \caption{Guidance and rules for generating entity-subentity vocabulary sets.}
    \label{fig:vocabs_helper}
\end{figure*}

\begin{figure*}[t]
    \centering
    \includegraphics[width=.98\linewidth]{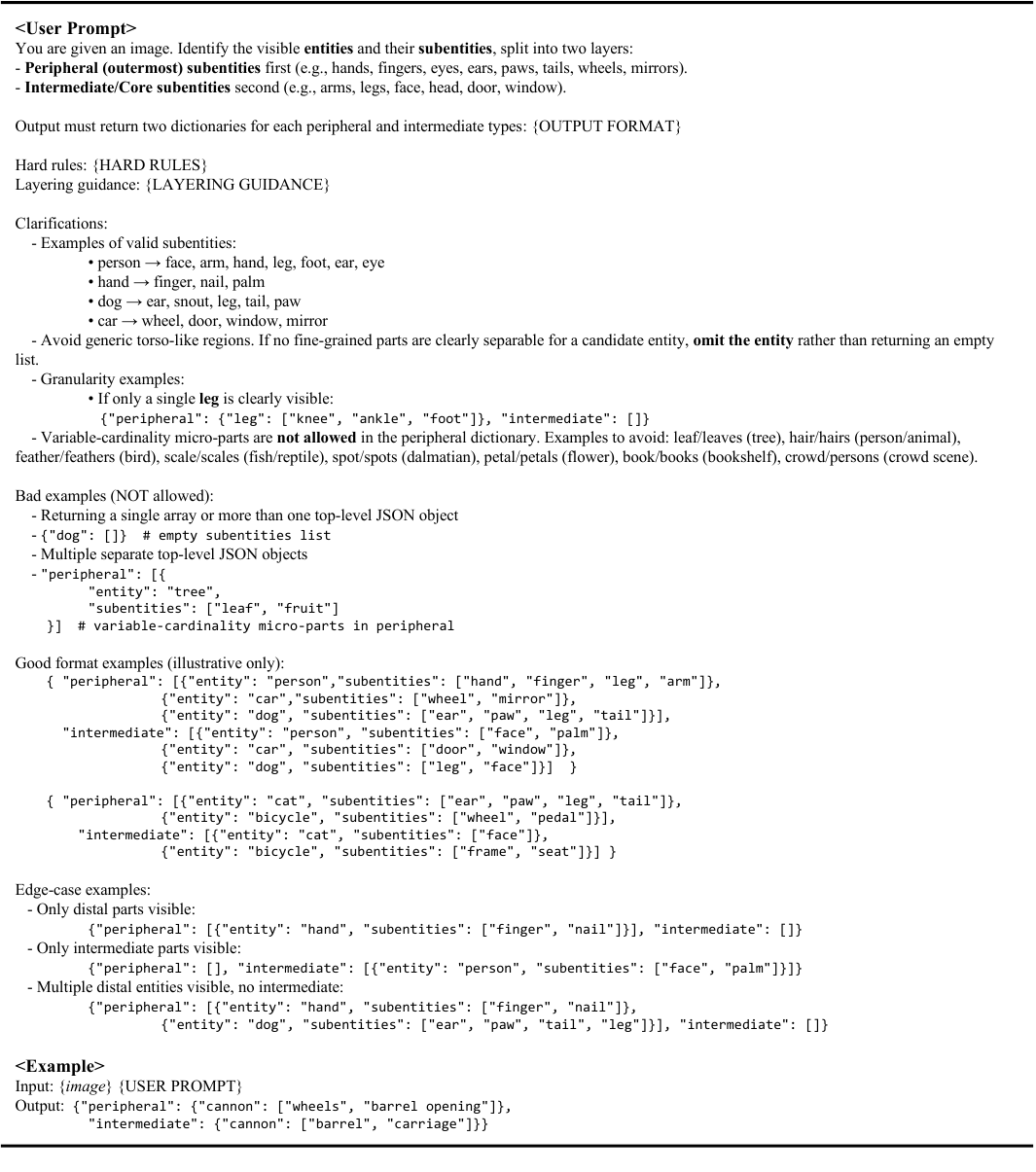}
    \caption{Full prompt for generating entity-subentity vocabulary sets.}
    \label{fig:vocabs}
\end{figure*}

\begin{figure}[t]
    \centering
    \includegraphics[width=\linewidth]{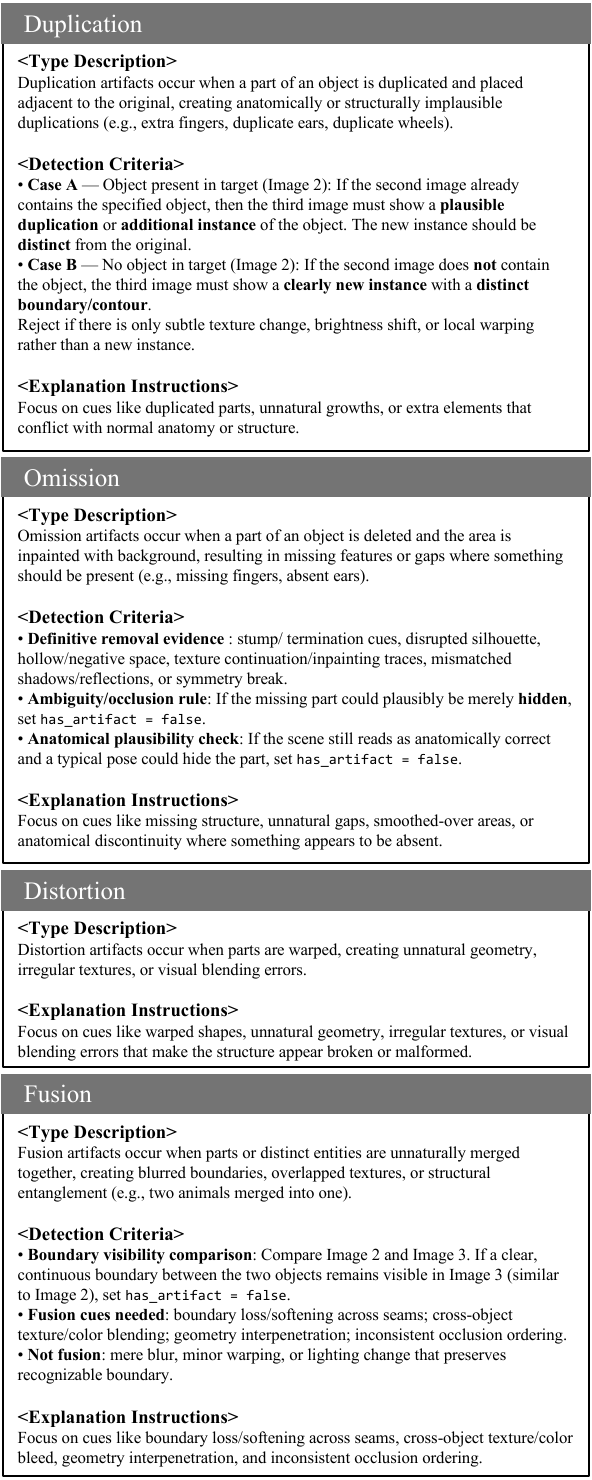}
    \vspace{-0.7cm}
    \caption{Type descriptions and instructions by artifact type.}
    \label{fig:artifact_types}
    \vspace{-0.5cm}
\end{figure}

\begin{figure}
    \centering
    \includegraphics[width=.98\linewidth]{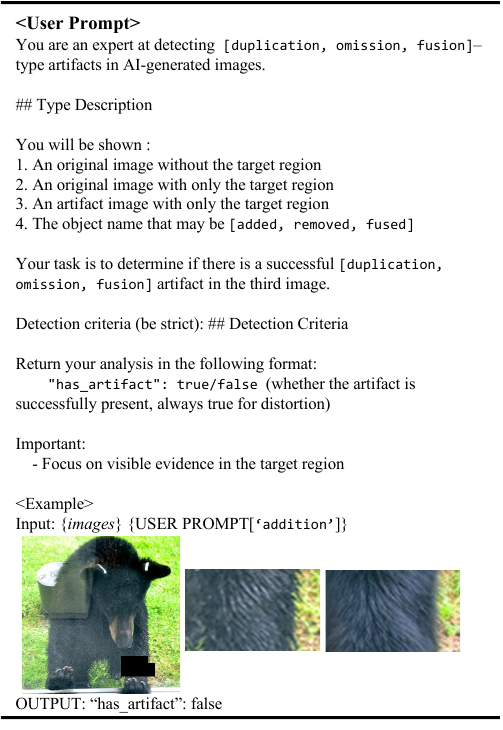}
    \vspace{-0.3cm}
    \caption{Full prompt and example for data filtering.}
    \label{fig:detection}
\end{figure}

\begin{figure}[t]
    \centering
    \includegraphics[width=.98\linewidth]{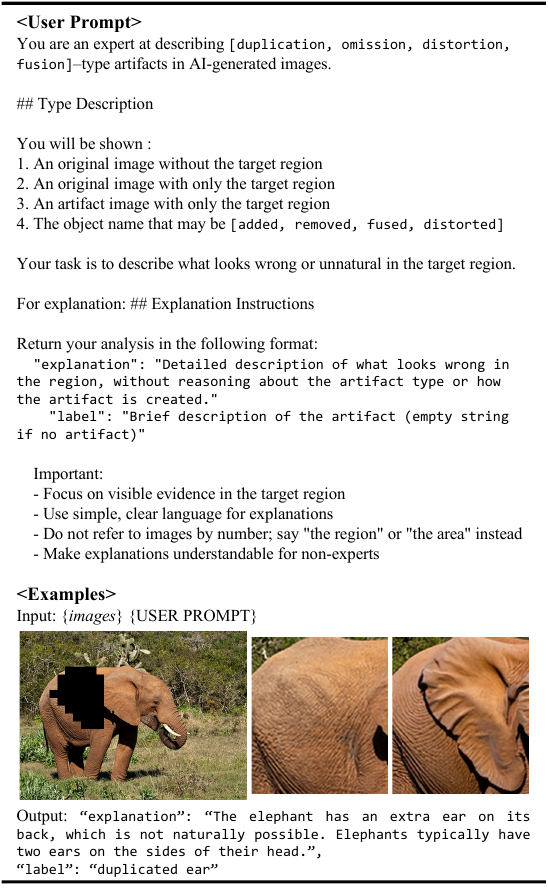}
    \vspace{-0.3cm}
    \caption{Full prompt and example for regional explanation generation.}
    \label{fig:expl_regional}
\end{figure}

\begin{figure}[t]
    \centering
    \includegraphics[width=.98\linewidth]{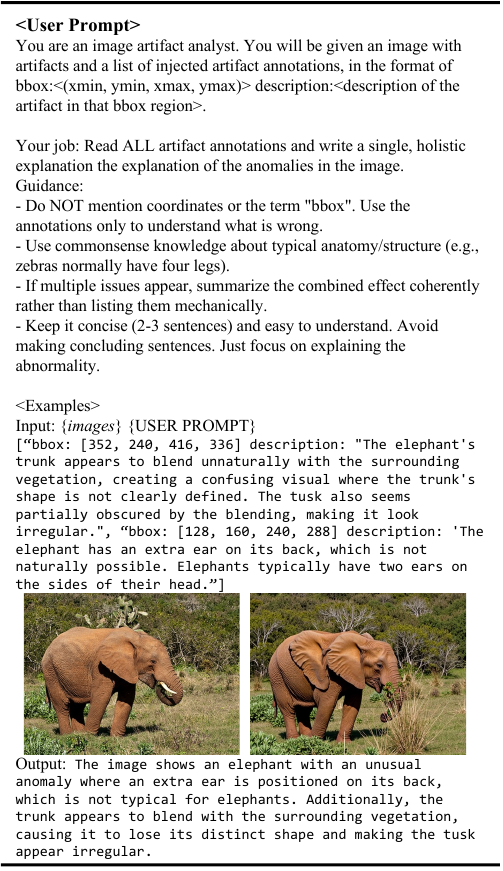}
    \vspace{-0.3cm}
    \caption{Full prompt and example for global explanation generation.}
    \label{fig:expl_global}
\end{figure}

Figure~\ref{fig:artiagent_viz} and Figure \ref{fig:final_results} illustrate the sampled images generated by the \algname{} pipeline and its synthesized annotations. The bounding boxes highlight the target patch area where artifacts are injected by the synthesis agent.

\subsection{Ablation Studies}
\label{apdx:ablations}
\begin{figure*}[t]
\centering
\includegraphics[width=.98\linewidth]{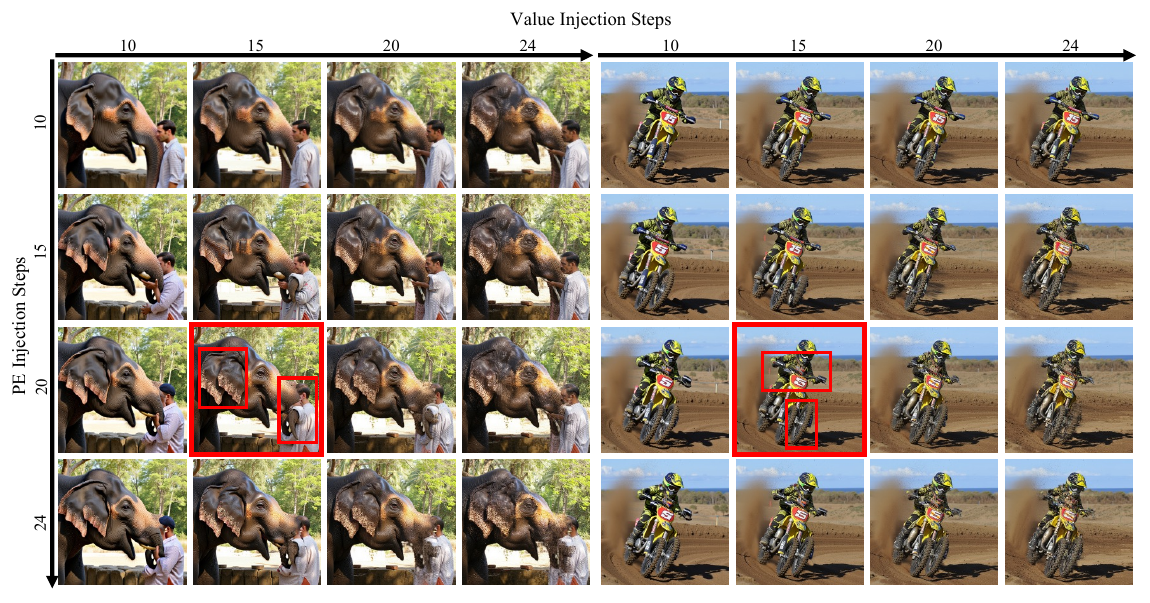}
\vspace{-0.4cm}
\caption{Hyperparameter study on PE injection and value injection steps (1). The image in the red box shows our selected configuration.}
\label{fig:ablations}
\end{figure*}

To justify the selected configuration of the synthesis agent in Appendix \ref{apdx:impl_details}, we conducted ablations on the PE injection steps along with the value injection blocks and visualized their results.
Figure~\ref{fig:ablations} shows how the choice of injection steps and value-injecting blocks affects artifact injection quality.

%
To investigate the effect of injection strength on artifact synthesis, we conducted an ablation study on the number of injection steps during the denoising process. Table~\ref{tbl:reb_ablation} shows that VLM performance peaks when artifacts are injected for 15 steps. Downstream performance degrades if the steps are too few (due to failed artifact injection) or too many (due to overall image quality degradation), supporting our choice of 15 injection steps.

\begin{table}[h]
\centering
\caption{\textbf{Ablation study on injection steps.} Performance of Qwen2.5-VL-7B on ArtiBench binary detection when trained with 1K ArtiAgent samples generated using varying numbers of injection steps out of 25 total steps.}
\label{tbl:reb_ablation}
\small
\begin{tabular}{c|cc}
\toprule
Steps & Acc & F1 \\
\midrule
5/25  & 0.513 & 0.377\\
10/25 & 0.583 & 0.565\\
15/25 & 0.586 & 0.570\\
20/25 & 0.540 & 0.477\\
\bottomrule
\end{tabular}
\end{table}



\begin{algorithm*}[p]
\caption{Add Tool}
\label{alg:add}
\footnotesize
\begin{algorithmic}[1]

\Require Reference patches $\mathcal{P}_R$, patch grid $(h_p, w_p)$, same-entity foreground $\mathcal{P}_{\text{ent}}$, same-subentity foreground $\mathcal{P}_{\text{sub}}$, ring thickness $\alpha$, distance weight $\lambda_{\text{dist}}$
\Ensure target-reference patch mapping $\mathcal{M} = \{(p_t, p_r)\}$

\State $(c_i, c_j) \gets \textsc{Centroid}(\mathcal{P}_R)$ \textcolor{blue}{\textit{/* subentity center in patch space */}}
\State $\mathcal{P}_{\text{ring}} \gets \textsc{PerimeterBand}(\mathcal{P}_R, (c_i,c_j), \alpha, h_p, w_p)$ \textcolor{blue}{\textit{/* candidate centroid patches band around the subentity */}}
\State $S \gets \emptyset$

\For{each $(i,j) \in \mathcal{P}_{\text{ring}}$}
    \State $\Delta_i \gets i - c_i,\ \Delta_j \gets j - c_j$
    \State $\mathcal{P}_{\text{shift}} \gets \{(r_i+\Delta_i, r_j+\Delta_j) : (r_i,r_j)\in\mathcal{P}_R\}$
    \State $r_{\text{self}} \gets \frac{|\mathcal{P}_{\text{shift}} \cap \mathcal{P}_R|}{|\mathcal{P}_{\text{shift}}|}$
    \State $r_{\text{ent}} \gets \frac{|\mathcal{P}_{\text{shift}} \cap (\mathcal{P}_{\text{ent}}\setminus\mathcal{P}_R)|}{|\mathcal{P}_{\text{shift}}|}$
    \State $r_{\text{sub}} \gets \frac{|\mathcal{P}_{\text{shift}} \cap \mathcal{P}_{\text{sub}}|}{|\mathcal{P}_{\text{shift}}|}$
    \State $d_{\text{L1}} \gets |i - c_i| + |j - c_j|$
    \State $g_{\text{dist}} \gets \frac{1}{1 + \lambda_{\text{dist}} d_{\text{L1}}}$
    \State $S(i,j) \gets (3 - r_{\text{self}} - r_{\text{ent}} - r_{\text{sub}}) \cdot g_{\text{dist}}$
\EndFor

\State $(i^\star, j^\star) \gets \arg\max_{(i,j)\in\mathcal{P}_{\text{ring}}} S(i,j)$
\State $\Delta_i^\star \gets i^\star - c_i,\ \Delta_j^\star \gets j^\star - c_j$
\State $\mathcal{M} \gets \{((r_i+\Delta_i^\star, r_j+\Delta_j^\star),(r_i,r_j)) : (r_i,r_j)\in\mathcal{P}_R\}$
\State \textbf{return} $\mathcal{M}$

\Statex
\hrulefill

\Function{Centroid}{$\mathcal{P}_R$}
    \State $n \gets |\mathcal{P}_R|$
    \State $c_i \gets \frac{1}{n} \sum_{(r_i,r_j)\in\mathcal{P}_R} r_i$
    \State $c_j \gets \frac{1}{n} \sum_{(r_i,r_j)\in\mathcal{P}_R} r_j$
    \State $c_i \gets \mathrm{round}(c_i)$,\quad $c_j \gets \mathrm{round}(c_j)$
    \State \Return $(c_i, c_j)$
\EndFunction

\Statex
\hrulefill

\Function{PerimeterBand}{$\mathcal{P}_R, (c_i,c_j), \alpha, h_p, w_p$}
    \State $\mathcal{P}_{\text{ring}} \gets \emptyset$
    \For{$i \gets 0$ to $h_p - 1$}
        \For{$j \gets 0$ to $w_p - 1$}
            \State $d_{\text{L1}} \gets |i - c_i| + |j - c_j|$
            \If{$1 \leq d_{\text{L1}} \leq \alpha$}
                \State $\mathcal{P}_{\text{ring}} \gets \mathcal{P}_{\text{ring}} \cup \{(i,j)\}$
            \EndIf
        \EndFor
    \EndFor
    \State \Return $\mathcal{P}_{\text{ring}}$
\EndFunction

\end{algorithmic}
\end{algorithm*}

\begin{algorithm*}[p]
\caption{Remove Tool}
\label{alg:remove}
\footnotesize
\begin{algorithmic}[1]

\Require Subentity patch set $\mathcal{P}_T$, patch grid $(h_p, w_p)$, 
same-entity foreground $\mathcal{P}_{\text{ent}}$, 
same-subentity foreground $\mathcal{P}_{\text{sub}}$, 
neighborhood radius $R$
\Ensure Target–reference mapping $\mathcal{M}$

\State $\mathcal{P}_{\text{nbr}} \gets \textsc{LocalNeighborhood}(\mathcal{P}_T, R, h_p, w_p)$ \textcolor{blue}{\textit{/* collect nearby non-target patches in the grid */}}

\State $\mathcal{P}_{\text{nbr-no-sub}} \gets \mathcal{P}_{\text{nbr}} \setminus \mathcal{P}_{\text{sub}}$
\State $\mathcal{P}_{\text{nbr-non-ent}} \gets \mathcal{P}_{\text{nbr}} \setminus \mathcal{P}_{\text{ent}}$

\If{$|\mathcal{P}_{\text{nbr-non-ent}}| > \tfrac{1}{2}|\mathcal{P}_{\text{nbr-no-sub}}|$}
    \State $\mathcal{P}_R^{\text{pool}} \gets \mathcal{P}_{\text{nbr-non-ent}}$
\Else
    \State $\mathcal{P}_R^{\text{pool}} \gets \mathcal{P}_{\text{nbr-no-sub}}$
\EndIf

\State $\mathcal{M} \gets \emptyset$
\For{each $p_t \in \mathcal{P}_T$}
    \State $p_r \gets \arg\min_{q \in \mathcal{P}_R^{\text{pool}}} \|p_t - q\|_1$
    \State $\mathcal{M} \gets \mathcal{M} \cup \{(p_t, p_r)\}$
\EndFor
\State \Return $\mathcal{M}$

\Statex \hrulefill

\Function{LocalNeighborhood}{$\mathcal{P}_T, R, h_p, w_p$}
    \State $\mathcal{P}_{\text{nbr}} \gets \emptyset$
    \For{each $(t_i,t_j) \in \mathcal{P}_T$}
        \For{$dy = -R$ to $R$}
            \For{$dx = -R$ to $R$}
                \If{$|dy| + |dx| \le R$ and $(dy,dx)\neq(0,0)$}
                    \State $p_i \gets t_i + dy$, $p_j \gets t_j + dx$
                    \If{$0 \le p_i < h_p$ and $0 \le p_j < w_p$ and $(p_i,p_j)\notin \mathcal{P}_T$}
                        \State $\mathcal{P}_{\text{nbr}} \gets \mathcal{P}_{\text{nbr}} \cup \{(p_i,p_j)\}$
                    \EndIf
                \EndIf
            \EndFor
        \EndFor
    \EndFor
    \State \Return $\mathcal{P}_{\text{nbr}}$
\EndFunction

\end{algorithmic}
\end{algorithm*}

\begin{algorithm*}[p]
\caption{Distort Tool}
\label{alg:distort}
\footnotesize
\begin{algorithmic}[1]

\Require Subentity patches $\mathcal{P}_T$, patch grid $(h_p,w_p)$, same-entity patches $\mathcal{P}_{\text{ent}}$, kernel type $k \in \{\text{shuffle}, \text{jitter}, \text{strip}\}$
\Ensure target-reference patch mapping $\mathcal{M} = \{(p_t, p_r)\}$

\If{$k = \text{shuffle}$}
    \State $\mathcal{P}_R \gets \textsc{ShuffleKernel}(\mathcal{P}_T)$ \textcolor{blue}{\textit{/* randomly permute subentity patches */}}
\ElsIf{$k = \text{jitter}$}
    \State $\mathcal{P}_R \gets \textsc{GaussianJitterKernel}(\mathcal{P}_T, \sigma, h_p, w_p, \mathcal{P}_{\text{ent}})$ \textcolor{blue}{\textit{/* locally jitter patches within the entity */}}
\ElsIf{$k = \text{strip}$}
    \State $\mathcal{P}_R \gets \textsc{StripShiftingKernel}(\mathcal{P}_T, S, h_p, w_p)$ \textcolor{blue}{\textit{/* shift patches along strips of the subentity */}}
\EndIf

\State $\mathcal{M} \gets \{(\mathcal{P}_T[i], \mathcal{P}_R[i]) : i = 1,\ldots,|\mathcal{P}_T|\}$
\State \textbf{return} $\mathcal{M}$

\Statex
\hrulefill

\Function{ShuffleKernel}{$\mathcal{P}_T$}
    \State $\mathcal{P}_R \gets \mathcal{P}_T$
    \State \textsc{RandomShuffle}$(\mathcal{P}_R)$
    \State \Return $\mathcal{P}_R$
\EndFunction

\Statex
\hrulefill

\Function{GaussianJitterKernel}{$\mathcal{P}_T, \sigma, h_p, w_p, \mathcal{P}_{\text{ent}}$}
    \State $\mathcal{P}_R \gets \emptyset$
    \For{each $(p_y, p_x) \in \mathcal{P}_T$}
        \State $\text{found} \gets \text{False}$
        \For{$a \gets 1$ to $A_{\max}$}
            \State $\delta_y \sim \mathcal{N}(0, \sigma^2)$,\quad $\delta_x \sim \mathcal{N}(0, \sigma^2)$
            \State $n_y \gets \mathrm{clip}(\mathrm{round}(p_y + \delta_y), 0, h_p - 1)$
            \State $n_x \gets \mathrm{clip}(\mathrm{round}(p_x + \delta_x), 0, w_p - 1)$
            \If{$\mathcal{P}_{\text{ent}} = \emptyset$ \textbf{or} $(n_y,n_x) \in \mathcal{P}_{\text{ent}}$}
                \State $\mathcal{P}_R \gets \mathcal{P}_R \cup \{(n_y,n_x)\}$
                \State $\text{found} \gets \text{True}$
                \State \textbf{break}
            \EndIf
        \EndFor
        \If{\textbf{not} found}
            \State $(n_y,n_x) \gets \textsc{NearestForegroundOrSelf}(p_y, p_x, \mathcal{P}_{\text{ent}})$
            \State $\mathcal{P}_R \gets \mathcal{P}_R \cup \{(n_y,n_x)\}$
        \EndIf
    \EndFor
    \State \Return $\mathcal{P}_R$
\EndFunction

\Statex
\hrulefill

\Function{StripShiftingKernel}{$\mathcal{P}_T, S, h_p, w_p$}
    \State Compute bounding box of $\mathcal{P}_T$ in patch space
    \State Determine direction $d \in \{\text{vertical},\text{horizontal}\}$ from aspect ratio
    \State Partition $\mathcal{P}_T$ into $S$ strips $\{\mathcal{S}_1,\dots,\mathcal{S}_S\}$ along $d$
    \State For each strip $\mathcal{S}_s$, sort patches to obtain an order $(p^{(s)}_1,\dots,p^{(s)}_{n_s})$
    \State Choose integer strip shifts $\{\Delta_s\}_{s=1}^S$ (alternating signs, increasing magnitudes)
    \State $\mathcal{P}_R \gets$ list of length $|\mathcal{P}_T|$
    \For{$s \gets 1$ to $S$}
        \State $\Delta \gets \Delta_s$ converted to patch units (circular shift)
        \For{$u \gets 1$ to $n_s$}
            \State $v \gets 1 + ((u + \Delta - 1) \bmod n_s)$
            \State Set reference of $p^{(s)}_u$ to $p^{(s)}_v$ in $\mathcal{P}_R$
        \EndFor
    \EndFor
    \State \Return $\mathcal{P}_R$
\EndFunction

\end{algorithmic}
\end{algorithm*}

\begin{algorithm*}[p]
\caption{Fuse Tool (Main)}
\label{alg:fuse-main}
\footnotesize
\begin{algorithmic}[1]

\Require Entity A patches $\mathcal{P}_A$, entity B patches $\mathcal{P}_B$, patch grid $(h_p,w_p)$, band radius $R$, max offset $R_{\text{off}}$, number of seeds $K$
\Ensure target-reference patch mapping $\mathcal{M} = \{(p_t, p_r)\}$

\State $\mathcal{P}_{\text{overlap}} \gets \mathcal{P}_A \cap \mathcal{P}_B$
\If{$\mathcal{P}_{\text{overlap}} = \emptyset$}
    \State \Return $\emptyset$
\EndIf

\State $\mathcal{P}_{\text{fg}} \gets \mathcal{P}_A \cup \mathcal{P}_B$
\State $\mathcal{P}_{A\setminus B} \gets \mathcal{P}_A \setminus \mathcal{P}_{\text{overlap}}$,\quad
       $\mathcal{P}_{B\setminus A} \gets \mathcal{P}_B \setminus \mathcal{P}_{\text{overlap}}$

\State $\mathcal{P}_T \gets \textsc{OverlapFusionBand}(\mathcal{P}_{\text{overlap}}, \mathcal{P}_{\text{fg}}, R, h_p, w_p)$ \textcolor{blue}{\textit{/* build foreground band around the overlap */}}
\If{$\mathcal{P}_T = \emptyset$}
    \State \Return $\emptyset$
\EndIf

\State $\mathcal{S} \gets \textsc{FarthestPointSampling}(\mathcal{P}_T, K)$ \textcolor{blue}{\textit{/* choose seeds that cover the band */}}
\State Initialize $\{\mathcal{R}_s\}_{s \in \mathcal{S}}$ as empty sets

\For{each $p \in \mathcal{P}_T$}
    \State $s^\star \gets \arg\min_{s \in \mathcal{S}} \|p - s\|_1$
    \State $\mathcal{R}_{s^\star} \gets \mathcal{R}_{s^\star} \cup \{p\}$
\EndFor

\State $\Omega \gets \{(\Delta_i,\Delta_j) : 1 \leq |\Delta_i| + |\Delta_j| \leq R_{\text{off}}\}$

\State $\mathcal{M} \gets \emptyset$
\For{each seed $s \in \mathcal{S}$}
    \State $\mathcal{R} \gets \mathcal{R}_s$
    \If{$\mathcal{R} = \emptyset$}
        \State \textbf{continue}
    \EndIf

    \State $\mathcal{P}_{\text{opp}} \gets \textsc{OppositeRegion}(s, \mathcal{P}_{A\setminus B}, \mathcal{P}_{B\setminus A}, \mathcal{P}_{\text{fg}}, \mathcal{P}_T)$ \textcolor{blue}{\textit{/* decide which side to fuse from */}}
    \State $\Delta^\star \gets \textsc{BestOffset}(\mathcal{R}, \mathcal{P}_{\text{opp}}, \Omega, h_p, w_p, \mathcal{P}_T)$ \textcolor{blue}{\textit{/* find best shared shift for this region */}}

    \For{each $p \in \mathcal{R}$}
        \State $p_r \gets \textsc{OffsetOrNearest}(p, \Delta^\star, \mathcal{P}_{\text{opp}}, h_p, w_p, \mathcal{P}_T)$ \textcolor{blue}{\textit{/* apply offset or nearest opposite patch */}}
        \State $\mathcal{M} \gets \mathcal{M} \cup \{(p, p_r)\}$
    \EndFor
\EndFor

\State \Return $\mathcal{M}$

\end{algorithmic}
\end{algorithm*}

\begin{algorithm*}[p]
\caption{Fuse Tool (Helpers)}
\label{alg:fuse-helpers}
\footnotesize
\begin{algorithmic}[1]

\Function{OverlapFusionBand}{$\mathcal{P}_{\text{overlap}}, \mathcal{P}_{\text{fg}}, R, h_p, w_p$}
    \State $\mathcal{P}_T \gets \emptyset$
    \For{each $(o_i,o_j) \in \mathcal{P}_{\text{overlap}}$}
        \For{$dy \gets -R$ to $R$}
            \For{$dx \gets -R$ to $R$}
                \If{$|dy| + |dx| \leq R$}
                    \State $v_i \gets o_i + dy$,\quad $v_j \gets o_j + dx$
                    \If{$0 \leq v_i < h_p$ \textbf{and} $0 \leq v_j < w_p$ \textbf{and} $(v_i,v_j) \in \mathcal{P}_{\text{fg}}$}
                        \State $\mathcal{P}_T \gets \mathcal{P}_T \cup \{(v_i,v_j)\}$
                    \EndIf
                \EndIf
            \EndFor
        \EndFor
    \EndFor
    \State \Return $\mathcal{P}_T$
\EndFunction

\Statex
\hrulefill

\Function{FarthestPointSampling}{$\mathcal{P}, K$}
    \If{$\mathcal{P} = \emptyset$}
        \State \Return $\emptyset$
    \EndIf
    \State $K \gets \min(K, |\mathcal{P}|)$
    \State Choose initial seed $s_1 \in \mathcal{P}$ (e.g., closest to centroid)
    \State $\mathcal{S} \gets \{s_1\}$; \quad $d(p) \gets \|p - s_1\|_1$ for all $p \in \mathcal{P}$
    \For{$m \gets 2$ to $K$}
        \State $s_m \gets \arg\max_{p \in \mathcal{P}} d(p)$
        \State $\mathcal{S} \gets \mathcal{S} \cup \{s_m\}$
        \For{each $p \in \mathcal{P}$}
            \State $d(p) \gets \min\{d(p), \|p - s_m\|_1\}$
        \EndFor
    \EndFor
    \State \Return $\mathcal{S}$
\EndFunction

\Statex
\hrulefill

\Function{OppositeRegion}{$s, \mathcal{P}_{A\setminus B}, \mathcal{P}_{B\setminus A}, \mathcal{P}_{\text{fg}}, \mathcal{P}_T$}
    \State $d_A \gets +\infty$,\quad $d_B \gets +\infty$
    \If{$\mathcal{P}_{A\setminus B} \neq \emptyset$}
        \State $d_A \gets \min_{a \in \mathcal{P}_{A\setminus B}} \|s - a\|_1$
    \EndIf
    \If{$\mathcal{P}_{B\setminus A} \neq \emptyset$}
        \State $d_B \gets \min_{b \in \mathcal{P}_{B\setminus A}} \|s - b\|_1$
    \EndIf
    \If{$d_A < d_B$}
        \State \Return $\mathcal{P}_{B\setminus A}$
    \ElsIf{$d_B < d_A$}
        \State \Return $\mathcal{P}_{A\setminus B}$
    \Else
        \State \Return $\mathcal{P}_{\text{fg}} \setminus \mathcal{P}_T$
    \EndIf
\EndFunction

\Statex
\hrulefill

\Function{BestOffset}{$\mathcal{R}, \mathcal{P}_{\text{opp}}, \Omega, h_p, w_p, \mathcal{P}_T$}
    \State $\Delta^\star \gets \text{None}$,\quad $m^\star \gets 0$
    \For{each $(\Delta_i,\Delta_j) \in \Omega$}
        \State $\text{count} \gets 0$
        \For{each $(v_i,v_j) \in \mathcal{R}$}
            \State $r_i \gets v_i + \Delta_i$,\quad $r_j \gets v_j + \Delta_j$
            \If{$0 \leq r_i < h_p$ \textbf{and} $0 \leq r_j < w_p$ \textbf{and} $(r_i,r_j) \in \mathcal{P}_{\text{opp}}$ \textbf{and} $(r_i,r_j) \notin \mathcal{P}_T$}
                \State $\text{count} \gets \text{count} + 1$
            \EndIf
        \EndFor
        \If{$\text{count} > m^\star$}
            \State $m^\star \gets \text{count}$,\quad $\Delta^\star \gets (\Delta_i,\Delta_j)$
        \EndIf
    \EndFor
    \State \Return $\Delta^\star$
\EndFunction

\Statex
\hrulefill

\Function{OffsetOrNearest}{$p, \Delta^\star, \mathcal{P}_{\text{opp}}, h_p, w_p, \mathcal{P}_T$}
    \State $(v_i,v_j) \gets p$
    \If{$\Delta^\star \neq \text{None}$}
        \State $(\Delta_i,\Delta_j) \gets \Delta^\star$
        \State $r_i \gets v_i + \Delta_i$,\quad $r_j \gets v_j + \Delta_j$
        \If{$0 \leq r_i < h_p$ \textbf{and} $0 \leq r_j < w_p$ \textbf{and} $(r_i,r_j) \in \mathcal{P}_{\text{opp}}$ \textbf{and} $(r_i,r_j) \notin \mathcal{P}_T$}
            \State \Return $(r_i,r_j)$
        \EndIf
    \EndIf
    \State \Return $\arg\min_{u \in \mathcal{P}_{\text{opp}}} \|(v_i,v_j) - u\|_1$
\EndFunction

\end{algorithmic}
\end{algorithm*}

\clearpage

\section{Benchmarks and Evaluation Protocols (\S~\ref{sec:benchmark} \& \S~\ref{subsec:vlm_eval})}

\subsection{Benchmark Datasets (\S~\ref{sec:benchmark})}
\label{apdx:eval_source}
\begin{table*}[t]
\small
\setlength{\tabcolsep}{4pt}
\caption{\textbf{Comparison of artifact benchmark datasets, their generative sources, and evaluation tasks.} 
\colorbox{lightgray}{Highlighted} entries denote dataset sources that were reused by subsequent benchmarks.}
\vspace{-0.3cm}
\centering
\begin{tabular}{@{}l l c c c c@{}}
\toprule
\textbf{Benchmark} 
& \textbf{Generative Sources} 
& \textbf{Sample} 
& \textbf{Bin.} 
& \textbf{Loc.} 
& \textbf{Exp.} \\
\midrule
RichHF~\citep{liang2024rich} 
  & SD2.1~\citep{rombach2022high}, SDXL~\citep{podell2024sdxl}, Dreamlike Photoreal~\citep{dreamlike2023}
  & 955 &  & \checkmark &  \\
\midrule
LOKI~\citep{ye2024loki} 
  & FLUX~\citep{FLUX}, SD1.4--2.1~\citep{rombach2022high}, Midjourney~\citep{midjourney2022}, 
    StyleGAN~\citep{karras2020stylegan2}, pix2pix~\citep{isola2017pix2pix}, CUT~\citep{park2020cut}
  & 229 &  & \checkmark & \checkmark \\
\midrule
SynthScars~\citep{kang2025legion} 
  & \colorbox{lightgray}{RichHF}~\citep{liang2024rich}, 
    \colorbox{lightgray}{Chameleon}~\citep{yan2025chameleon}, 
    Midjourney~\citep{midjourney2022}, 
    DALL{\textperiodcentered}E 3~\citep{openai2023dalle3}, 
    SD1.x~\citep{rombach2022high}
  & 1K &  & \checkmark & \checkmark \\
\midrule
\benchname{}
  & SD3.5~\citep{esser2024sd3}, FLUX~\citep{FLUX}, Qwen-Image~\citep{wu2025qwenimage}, Nano-Banana~\citep{google2025nanobanana}
  & 1K & \checkmark & \checkmark & \checkmark \\
\bottomrule
\end{tabular}
\vspace{-0.1cm}
\label{apdx_tbl:eval_source}
\end{table*}
Table \ref{apdx_tbl:eval_source} shows the fully cited artifact benchmark datasets and the sources used for generation. Details on the dataset's metadata formats for artifact region representations and image explanations are elaborated below. 
\begin{itemize}[leftmargin=10pt,noitemsep]
    \item\textbf{RichHF.} The RichHF dataset, with 995 images sampled and annotated from the Pick-a-Pic dataset, provides the artifact map in a heatmap format, highlighting the probability of abnormal regions in the image. Annotations and scores are generated by the trained multi-modal transformer to predict human perception-aligned feedback.  
    
    \item\textbf{LOKI.} The LOKI benchmark is a multi-modal synthetic-data detection dataset, covering image, video, text, audio, and 3D content, and comprises roughly 18,000 curated questions across 26 subcategories. It includes coarse real vs. synthetic judgments, multiple-choice detection, anomaly / artifact region selection, and explanation tasks. We use the image artifact subset of the dataset, which consists of 229 images. 
    
    \item\textbf{SynthScars.} The SynthScars dataset consists of 12,236 fully synthetic images across four distinct content types (Human, Object, Scene, Animal) and three artifact categories (Physics, Distortion, Structure). Each image is annotated with pixel-level segmentation masks delimiting artifact regions, detailed textual explanations of the artifact(s), and artifact-category labels.  
    
\end{itemize}

\begin{figure}
    \centering
    \includegraphics[width=.98\linewidth]{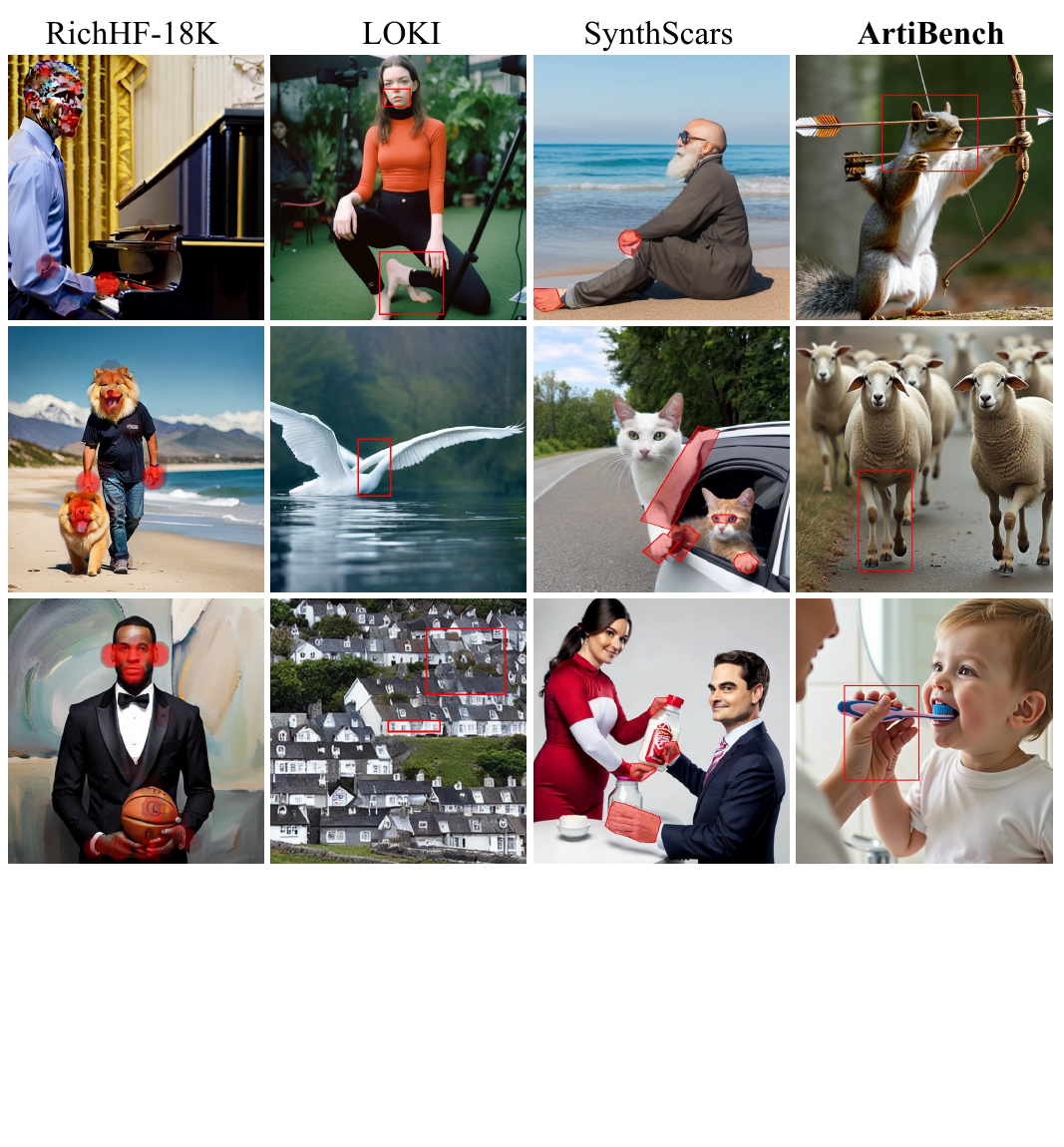}
    \vspace{-2cm}
    \caption{Artifact examples of the four benchmarks, listed in the order of publishment.}
    \label{fig:bench_viz}
\end{figure}

\myparagraph{Visualizations.}
Visualizations of the evaluated benchmarks, including \benchname{}, are shown in Figure~\ref{fig:bench_viz}. It is clearly visible that even though our benchmark dataset shows overall better quality of image generation compared to the previous benchmarks, image artifacts are still visible in the generated images. We emphasize the importance of our timely benchmark that successfully represents structural artifacts remaining in the most recent diffusion models.

\subsection{ArtiBench Generation (\S~\ref{sec:benchmark})}
\label{apdx:bench_gen}
\myparagraph{Image Generation.}
\benchname{} is built on a set of images generated by state-of-the-art diffusion models, reflecting the most timely artifacts appearing in current image generation models. We use five different models for image generation and three reliable prompt sources. After generating a set of images with the five models from randomly sampled prompts, the images were annotated according to the strictly guided annotation process.

\myparagraph{Annotation Pipeline.}
With the set of diffusion-generated images, we construct \textbf{\benchname{}} following a guided 4-step annotation process: (1) classification, (2) bounding box labeling, (3) explanation generation, and (4) expert curation.

\smallskip
\begin{itemize}[leftmargin=10pt,noitemsep]
    \item \textbf{Classification.} Annotators are asked to meticulously observe the images and determine whether there are any artifacts visible in the image. The caption used for T2I generation is provided with the image for better understanding of generative intentions and respecting the model's capability to follow instructions on generating abnormalities. All images classified as having artifacts proceed to the next step, where they are shuffled and redistributed to the annotators to mitigate human bias in the examination process.
    \item \textbf{Bounding Box Labeling.} For the artifact-containing images from the classification step, annotators are expected to generate bounding boxes, identify the type of artifact, and write a short description about what abnormalities exist in the specified region.
    \item \textbf{Explanation Generation.} The final annotations are used as the input of the VLM query to generate a comprehensive and polished explanation for the full image. With the prompt used for generation and the list of pairs of bounding box and captions as the input, the VLM is required to output a global explanation of the image regarding all artifacts mentioned in the captions.
    \item \textbf{Expert Curation.} With the set of images and the metadata completed, we curate the images to build the final version of our 1K benchmark. Images are carefully selected to ensure a balanced set that provides a clear and straightforward view of plausible artifacts.
\end{itemize}
\vfill

\subsection{Evaluation Protocol (\S~\ref{subsec:vlm_eval})}
\label{apdx:eval_process}
\subsubsection{Binary Classification}
\begin{itemize}
    \item \textbf{Accuracy.} The rate of true predictions is measured to show basic performance of accurate classification.
    \item \textbf{Macro F1.} In addition, we use the macro F1 score, to achieve fair comparison across unbalanced datasets and biased predictions. F1 scores are relatively calculated for positive and negative samples to be averaged, exposing the model's capability to precisely capture both artifact presence and absence without random guessing.
\end{itemize}

\subsubsection{Localization}
We alleviate the unfairness between divergent representations of artifact regions, including bounding boxes, polygonal segmentation maps, and heatmaps, by mapping all representations to pixel-wise binary maps.
\begin{itemize}[leftmargin=10pt,noitemsep]
    \item \textbf{IoU.} With the binary maps obtained, we calculate the pixel-wise IoU between the foreground region predictions and ground-truth areas. IoU scores prefer tight and highly overlapping predictions, providing intuitions on precise predictions but highly possible of penalizing bounding box representations for over-prediction.
    \item \textbf{F1.} Pixel-wise predictions are measured to capture true predictions on a fine-grained basis. Unlike the macro F1 score used in binary detection tasks, F1 scores are measured only for \textit{positive} ground truth regions to prioritize true detections. With each pixel prediction classified, we use the pixel count for true positive (TP), false positive (FP), and false negative (FN) predictions to use them for calculating the F1 score. This method lessens the penalty on loose region representations of bounding boxes over segmentation maps or heatmaps.
\end{itemize}

\subsubsection{Explanation}
\begin{itemize}
    \item \textbf{ROUGE-L.} The ROUGE-L score shows the proportion of the longest overlapping phrase among the full data. This captures the words or phrases that focus on specific artifact regions and objects, with higher scores showing that the model better understands the visual artifact.
    \item \textbf{CSS.} Cosine similarity was measured on sentence embeddings generated by sentence-transformers~\cite{reimers-2019-sentence-bert}. CSS portrays the general similarity in context between the descriptions regarding the full scenery's plausibility.
\end{itemize}

\subsubsection{Evaluation Prompt of VLMs}
\label{apdx_subsec:eval_prompt}
\begin{figure}[h]
    \centering
    \includegraphics[width=.98\linewidth]{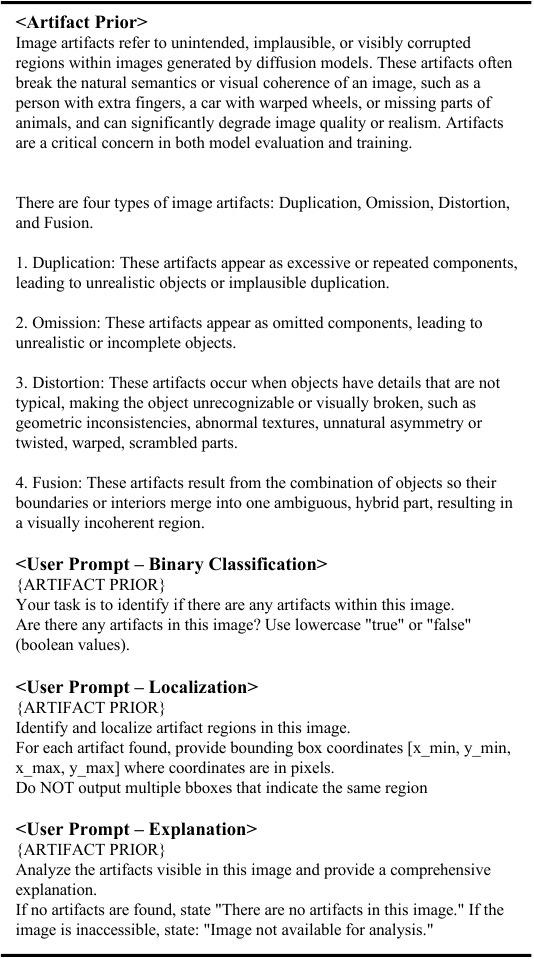}
    \vspace{-0.3cm}
    \caption{Evaluation prompts for VLMs.}
    \label{fig:eval_prompts}
\end{figure}

Figure~\ref{fig:eval_prompts} shows the specific prompts used for the evaluation on VLMs. Artifact priors are shared for all tasks, providing understanding of the structural artifacts we are focusing on.


\subsection{VQA Dataset Structure (\S~\ref{subsec:vlm_eval})}
\label{apdx:vlm_training}
This section presents the structure of our multi-turn visual question answering (VQA) dataset, which provides the supervision used to train VLMs in artifact detection, localization, and explanation. Each data instance is derived from a paired clean–artifact image generated by \algname{} with the synthesized annotations as in Figure~\ref{fig:final_results}, enabling the construction of tightly aligned conversations that elicit the model’s ability to identify normal content and reason about artifact regions.

\subsubsection{VQA Template}

Each ArtiAgent instance yields two conversations: one for the clean reconstruction image and one for the corresponding artifact image. Tables~\ref{apdx_tbl:vqa_clean} and \ref{apdx_tbl:vqa_artifact} summarize the question-answer templates implemented across these two settings.

\begin{table}[h]
\centering
\small
\begin{tabular}{l | l}
\toprule
\textbf{Task} & \textbf{Template} \\
\midrule

\multirow{2}{*}{Bin.} &
Q: Does this image contain any visual artifacts? \\
& A: No. \\[4pt]

\multirow{2}{*}{Loc.~1} &
Q: Locate the \textlangle entity\textrangle's \textlangle subentity\textrangle. \\
& A: \textlangle bbox\textrangle \\[4pt]

\multirow{2}{*}{Loc.~2} &
Q: Is there a \textlangle entity\textrangle's \textlangle subentity\textrangle\ in \textlangle bbox\textrangle? \\
& A: Yes / No \\[4pt]

\multirow{2}{*}{Exp.} &
Q: Describe the clean image. \\
& A: \textlangle image\_caption\textrangle \\
\bottomrule
\end{tabular}
\vspace{-0.2cm}
\caption{VQA templates for clean images.}
\label{apdx_tbl:vqa_clean}
\end{table}

\begin{table}[h]
\centering
\small
\begin{tabular}{l | l}
\toprule
\textbf{Task} & \textbf{Template} \\
\midrule

\multirow{2}{*}{Bin.} &
Q: Does this image contain any visual artifacts? \\
& A: Yes. \\[4pt]

\multirow{2}{*}{Loc.~1} &
Q: Provide bounding boxes for all artifact regions. \\
& A: [\textlangle bbox\textrangle] \\[4pt]

\multirow{2}{*}{Loc.~2} &
Q: Explain why region \textlangle bbox\textrangle\ is an artifact. \\
& A: \textlangle label\textrangle \\[4pt]

\multirow{2}{*}{Exp.} &
Q: Describe all artifacts in the image. \\
& A: \textlangle explanation\textrangle \\
\bottomrule
\end{tabular}
\vspace{-0.2cm}
\caption{VQA templates for artifact-injected images.}
\label{apdx_tbl:vqa_artifact}
\end{table}

\subsubsection{VLM Training Configuration}

We train the VLM using a two-stage supervised fine-tuning setup. We use Qwen2.5-VL-7B-Instruct~\cite{bai2025qwen2.5vl} and Intern3.5-VL-8B~\cite{zhu2025internvl3} as the backbone, and use the identical configurations. In the first stage, we fine-tune only the language model and multi-modal projector while keeping the vision encoder frozen, using a learning rate of $1{\times}10^{-5}$, batch size 64, cosine decay scheduling, and one training epoch. In the second stage, we unfreeze the vision encoder and continue training with a smaller learning rate of $1{\times}10^{-6}$, with 200 steps. Both stages use the same VQA dataset.

\section{Mitigating Artifacts in Diffusion Models (\S~\ref{subsubsec:reward_guided_gen} \& \S~\ref{subsubsec:artifact_correction})}

\subsection{Reward-guided generation (\S~\ref{subsubsec:reward_guided_gen})}
\label{apdx:reward_guided_gen}

\myparagraph{Verifier Training.}
We train an artifact verifier modeled with Bradley-Terry model~\cite{btmodel} to score images by how likely they are to be artifact-free. The model is trained to assign higher scores to the clean image and lower scores to the artifact-injected image. The verifier uses a frozen ViT-B/16 encoder with a lightweight MLP head, and is optimized using AdamW (learning rate $1\times10^{-3}$, weight decay $1\times10^{-4}$, L2 regularization $1\times10^{-4}$). Training uses a batch size of 32 for 5 epochs. Each batch additionally includes two cross-scene negative pairs to encourage content-agnostic ranking.

\myparagraph{Test-Time Scaling.}
At inference time, we use the verifier as a reward model in a compute-scaled best-of-$N$ sampling procedure~\cite{ma2025ttdiffusion}. For each prompt, round $r$ samples $2^{r}$ independent latent noises, generates all corresponding images, and evaluates them with the verifier. The highest-scoring image is retained, and the search space doubles in the next round. This random-search strategy expands the candidate pool exponentially, enabling the diffusion model to reliably discover images with fewer artifacts without modifying the weights of the model.

\subsection{Image Correction (\S~\ref{subsubsec:artifact_correction})}
\label{apdx:image_correction}

In our artifact-correction pipeline, all VLM interactions are carried out using the
Qwen2.5-VL-7B model fine-tuned with ArtiAgent supervision. This unified VLM is responsible
for localizing artifact regions, generating artifact-free captions, and verifying whether the
corrected content remains flawed. We describe each of the three prompts used in the loop below. The actual prompt is shown in Table~\ref{apdx_tab:image_correction_prompts}.

\begin{table}[h]
\small
\centering
\renewcommand{\arraystretch}{1.2}
\setlength{\tabcolsep}{4pt}

\begin{tabular}{p{1.5cm}| p{5.5cm}}
\toprule
\textbf{Prompt} & \textbf{Instruction Text} \\
\midrule

Localization
&
``Provide the bounding box for the artifact region.
Format as [x\_min, y\_min, x\_max, y\_max].
Only output the bounding box.'' 
\\[2pt]

Captioning&
``Describe what the image would look like if it had no artifacts.
Provide a short caption of the clean scene.'' 
\\[2pt]

Verifying&
``Is there an artifact within \textlangle B\textrangle?
Answer Yes or No.'' 
\\

\bottomrule
\end{tabular}
\vspace{-0.2cm}
\caption{Prompts used by the VLM in the artifact-correction loop.}
\label{apdx_tab:image_correction_prompts}
\end{table}
\begin{algorithm}[h]
\caption{VLM-Guided Artifact Correction Loop}
\label{alg:artifact_correction}
\begin{algorithmic}[1]

\Require Image $I$ containing at least one artifact

\State
{\color{blue} /* localize artifact region */ }

\State $B \gets \text{VLM}(I, p_{\text{localize}})$
\State
{\color{blue} /* generate caption of the image */ }

\State $c \gets \text{VLM}(I, p_{\text{caption}})$
\While{true}
    \State
    {\color{blue} /* Inpaint region $B$ using caption $c$ */ }

    \State $I \gets \text{Inpaint}(I, B, c)$

    \State
    {\color{blue} /* check artifact existence inside $B$ */ }

    \State $r \gets \text{VLM}(I, p_\text{verify})$

    \If{not $r$}        
    \State \Return $I$
    \EndIf

\EndWhile

\end{algorithmic}
\end{algorithm}

\myparagraph{Localization prompt $p_{\text{localize}}$.}
This prompt instructs the VLM to identify the spatial extent of the artifact within the input
image. The model is asked to produce a single bounding box formatted as
\texttt{[x\_min, y\_min, x\_max, y\_max]} without additional commentary. This strict output
format ensures deterministic parsing. The resulting bounding box $B$ is computed once at the
beginning of the procedure and kept fixed across all subsequent iterations, providing spatial
stability for the iterative refinement process.

\myparagraph{Captioning prompt $p_{\text{caption}}$.}
To guide inpainting, the VLM is prompted to describe the overall image without specifying the artifacts in the image. Rather than summarizing the corrupted content, the model is explicitly instructed to generate a short caption that reflects the intended clean version of the scene. This caption serves as the semantic conditioning signal for the FLUX inpainting pipeline~\cite{fluxinpaint},
which synthesizes corrected content inside the region $B$.

\myparagraph{Verifying prompt $p_{\text{verify}}$.}
After each inpainting step, the VLM is queried to assess whether the corrected region $B$
still contains an artifact. The prompt restricts the judgment to the localized region,
returning a binary response (\texttt{Yes} or \texttt{No}). This local verification prevents the
algorithm from drifting to unrelated image regions and directly determines whether the loop
terminates or continues.

Together, these three prompts coordinate the interaction between the ArtiAgent-trained VLM
and the FLUX inpainting pipeline: the VLM identifies the corrupted area, provides semantic
guidance for correction, and validates the result, while the inpainting model performs the
pixel-level repair. This iterative coupling enables consistent and stable reduction of visual
artifacts while preserving overall image semantics. The whole procedure is shown in Algorithm~\ref{alg:artifact_correction}.

\end{document}